\documentclass{ecai} 

\usepackage{latexsym}
\usepackage{amssymb}
\usepackage{amsmath}
\usepackage{amsthm}
\usepackage{booktabs}
\usepackage{enumitem}
\usepackage{graphicx}
\usepackage{color}

\usepackage{multirow}
\usepackage{microtype}

\usepackage{mathtools}

\DeclareMathOperator*{\argmin}{argmin}

\newcommand{\landl}{\mathbin{\landx{L}}}
\newcommand{\lands}{\mathbin{\landx{S}}}
\newcommand{\landp}{\mathbin{\landx{P}}}

\newcommand{\landx}[1]{%
  \ooalign{%
    $\land$\cr
    \noalign{\kern.5ex}
    \hidewidth$\scriptscriptstyle#1\mkern1.5mu$\hidewidth\cr
  }%
}
\newcommand{\lorx}[1]{%
  \ooalign{%
    $\lor$\cr
    \noalign{\kern.5ex}
    \hidewidth$\scriptscriptstyle#1\mkern1.5mu$\hidewidth\cr
  }%
}

\usepackage[capitalize]{cleveref}
\crefname{section}{Sec.}{Sec.}
\crefname{equation}{Eq.}{Eq.}
\crefname{definition}{Def.}{Def.}
\crefname{table}{Tab.}{Tab.}
\crefname{figure}{Fig.}{Fig.}
\crefname{appendix}{App.}{App.}
\creflabelformat{equation}{#2\textup{#1}#3}

\newtheorem{definition}{Definition}

\newcommand{\BibTeX}{B\kern-.05em{\sc i\kern-.025em b}\kern-.08em\TeX}

\begin{document}


\begin{frontmatter}


\paperid{1329} 


\title{Binarizing Physics-Inspired GNNs \\
for Combinatorial Optimization}


\author[A]{\fnms{Martin}~\snm{Krutsk\'y}\orcid{0009-0000-9710-1147}\thanks{Corresponding Authors. Emails: krutsma1@fel.cvut.cz, gustav.sir@cvut.cz}}
\author[A]{\fnms{Gustav}~\snm{{\v S}\'ir}\orcid{0000-0001-6964-4232}\footnotemark[*]}
\author[A]{\fnms{Vyacheslav}~\snm{Kungurtsev}\orcid{0000-0003-2229-8824}}
\author[B,A,C]{\fnms{Georgios}~\snm{Korpas}\orcid{0000-0003-3850-4979}} 


\address[A]{Department of Computer Science, Faculty of Electrical Engineering, Czech Technical University}
\address[B]{Quantum Technologies Group, HSBC Labs, Singapore}
\address[C]{Archimedes Research Unit on AI, Data Science and Algorithms, Athena Research Center}


\begin{abstract}
Physics-inspired graph neural networks (PI-GNNs) have
been utilized as an efficient unsupervised framework for relaxing combinatorial optimization problems encoded through a specific graph structure and loss, reflecting dependencies between the problem's variables.
While the framework has yielded promising results in various combinatorial problems, we show that the performance of PI-GNNs systematically plummets with an increasing density of the combinatorial problem graphs.
Our analysis reveals an interesting phase transition in the PI-GNNs' training dynamics, associated with degenerate solutions for the denser problems, highlighting a discrepancy between the relaxed, real-valued model outputs and the binary-valued problem solutions.
To address the discrepancy, we propose
principled alternatives to the naive strategy used in PI-GNNs by building on insights from fuzzy logic and binarized neural networks.
Our experiments demonstrate that the portfolio of proposed methods significantly improves the performance of PI-GNNs in increasingly dense settings.
\end{abstract}

\end{frontmatter}


\section{Introduction}
\label{sec:introduction}
With the advent of deep learning and its penetration into classical AI fields, such as automated planning and logical reasoning, combinatorial optimization (CO) has emerged as another promising area for its application. Among various deep learning approaches to CO, ranging from end-to-end machine learning algorithms to hybrid systems~\citep{cappart2023combinatorial}, physics-inspired graph neural networks (PI-GNNs)~\citep[][]{schuetz2022combinatorial} have gained particular attention due to their generality and interpretable physical intuition.
PI-GNNs leverage a specific loss function to encode the objective of an optimization problem, enabling unsupervised training to converge on solutions that minimize the objective.
This formulation is highly adaptable across various combinatorial optimization domains, establishing PI-GNNs as a promising general-purpose CO-solving framework.
The theoretical promise of PI-GNNs draws on insights from the overparameterization regime of neural networks, where multiple studies have shown that---under specific assumptions---the noise in
stochastic gradient descent leads to convergence to global minima~\citep{mei2018mean, zou2020gradient}. 
Coupled with their inherent parallelizability and computational power of modern GPUs, PI-GNNs emerge as a potentially more scalable and efficient alternative to classical solvers.

Despite their potential, PI-GNNs have recently faced critique for underperforming relative to efficient domain-specific heuristics on sparse graphs~\citep{angelini2023modern,boettcher2023inability}. In response, it has been argued that PI-GNNs exhibit superior performance on \textit{dense} graph instances~\citep{schuetz2023reply1}, considered better suited for the natively parallel deep learning-based approach~\citep{schuetz2023reply2}.
However, the experimental analysis has focused merely on sparse $d$-regular graphs ($d=3$, $d=5$),\footnote{with edge density~\citep{diestel2000graph} $\rho(\mathcal{G}) \leq 0.051$ across all experiments. \citet{schuetz2023reply2} claim that $7$ MaxCut problems from the Gset dataset~\citep{Ye2003} are an exception, but their density is low, as well: $\rho(\mathcal{G}) \leq 0.015$. Truly dense problems were tackled only within a different domain~\citep{schuetz2022graph}.} despite denser graphs having an essential status in domains involving network analysis~\citep{melancon2006just, nanavati2008analyzing}.

To investigate the setting, this work addresses the performance of PI-GNNs on graphs with progressively increasing density. Our results confirm favorable performance on the, originally-selected, sparse instances~\citep{schuetz2022combinatorial}; however, contrary to the previous claims~\citep{schuetz2023reply1,schuetz2023reply2}, we demonstrate that the performance of PI-GNNs consistently plummets as the density increases.
Our analysis of the training dynamics of PI-GNNs reveals an intriguing phenomenon: in sparse settings, the distribution of model output values undergoes a phase shift where all the values initially converge toward near-zero and, subsequently, for the selected nodes, toward near-one values. This behavior appears essential for the successful convergence of the original PI-GNNs; however, it rapidly diminishes as the graph density increases.

To address this phenomenon, we focus on the employed relaxation strategy of PI-GNNs, currently trained as standard deep learning models, with continuous outputs that are only thresholded after training to produce the binary solution for the underlying combinatorial problem.
In contrast, we propose to inform the training process about the binary nature of the underlying task through principled binarization techniques.
Specifically, using concepts from fuzzy logic~\citep{klir1995fuzzy} and binarized neural networks~\citep{hubara2016binarized}, we resolve the phase shift issues and significantly outperform the original PI-GNNs across graphs of varying connectivity, especially in denser settings.

\section{Background}
\label{sec:background}
We target physics-inspired (\cref{sec:spin}) GNN models (\cref{sec:gnns}) for solving combinatorial optimization problems (\cref{sec:co}).
\subsection{Combinatorial Optimization}
\label{sec:co}

Combinatorial optimization (CO) formalizes problems with a discrete solution space, making traditional efficient continuous optimization techniques, such as gradient-based methods, not directly applicable. 
\begin{definition}[Combinatorial Optimization Problem
]
\label{def:co}
An \emph{instance} of a \emph{combinatorial optimization problem} is a triplet $(\Omega, \mathcal{F}, c)$, where $\Omega$ is a finite set, $\mathcal{F} \subseteq 2^\Omega$ is the set of feasible solutions,
and  $c: 2^\Omega \to \mathbb{R}$ is a cost function.
The solution to such a problem is $S^*={\argmin}_{S \in \mathcal{F}}\ c(S)$.
\end{definition}
Solving combinatorial optimization problems thus requires selection of an optimal subset of a finite (or countable) domain set. Due to a combinatorial explosion, the number of candidates
can be enormous. Therefore, many combinatorial optimization problems are notoriously hard to solve, with no general algorithmic solutions that would run in polynomial time w.r.t. the input size~\citep{cook1971}.

\begin{figure}[t!]
    \centering
    \includegraphics[width=.9\columnwidth]{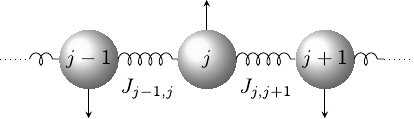}
    \caption{Schematic of an Ising spin chain in 1D, with arrow/spin \textit{down} corresponding to assignment $0$, and \textit{up} to $1$, respectively.}
    \label{fig:spin}
\end{figure}

\paragraph{Graph Representation in CO}
\label{sec:significance}
When~\citet{karp2010reducibility} introduced the hierarchy of $21$ combinatorial \textit{NP}-complete problems, a significant subset shared a common structure, being defined over graphs. These include problems of \textit{maximum cut}, \textit{maximum independent set}, \textit{maximum clique}, and \textit{graph partitioning}, among others. Moreover, Karp demonstrated that even the remaining problems in this hierarchy can be polynomially reduced to one of these graph-based problems.
Here we introduce two \textit{NP}-hard CO problems that will be used in the paper.

\begin{definition}[Maximum Cut]
\label{def:maxcut}
Given an undirected graph $\mathcal{G} = (\mathcal{V}, \mathcal{E})$, partition $\mathcal{V}$ into two disjoint sets $\mathcal{V}_1, \mathcal{V}_2$, such that the number of edges with one node in $\mathcal{V}_1$ and the other in $\mathcal{V}_2$ is maximal:
\begin{equation}
\max_{\mathcal{V}_1 \subseteq \mathcal{V}} |\{e\ |\ e = \{v_i, v_j\}, v_i \in \mathcal{V}_1 \Leftrightarrow v_j \notin \mathcal{V}_1\}|
\end{equation}
\end{definition}

\begin{definition}[Maximum Independent Set]
\label{def:mis}
Given an undirected graph $\mathcal{G} = (\mathcal{V}, \mathcal{E})$ and an indicator function $i: \mathcal{V} \rightarrow \{0, 1\}$, the maximum independent set is a subset of maximal size consisting of vertices that do not share an edge:
\begin{equation}
\max\limits_{V \subseteq \mathcal{V}} |\{v\ |\ i(v) = 1, \forall v^\prime \in \mathcal{V}:\{v, v^\prime\} \in \mathcal{E} \Rightarrow i(v^\prime) = 0\}|
\end{equation}
\end{definition}

\paragraph{Relaxation of CO Problems}
Due to their computational complexity, solving CO problems often involves tackling a \emph{relaxation}, where the discrete decision variables, indicating the solution subset, are replaced with continuous ones. Ideally, the solution to the relaxed problem is naturally integral, coinciding with the optimal solution of the original problem. However, in most practical scenarios, relaxation techniques provide solutions that either need to be appropriately rounded or act as intermediate solutions.
This fact plays a prominent role in CO approaches, such as in the Branch and Bound (B\&B) algorithm using solutions to linear program relaxations for search guidance~\citep{baum1981integer}.
Similarly, the Lovász relaxation~\citep{grotschel1995combinatorial} for approximating integer programs facilitates efficient optimization and rounding. The challenge of converting continuous relaxations into discrete decisions is thus central in both classical~\citep{berthold2014rens} and machine learning-based methods.

\subsection{Physics-Based Approaches to CO}
\label{sec:spin}

An elegant strategy for tackling CO problems is to encode the objective as the ``energy'' of a spin-glass system \citep{mezard2006spin,Yamamoto2017}. Spin-glass models originate in statistical mechanics and are used extensively to study magnetic crystals, superconductivity, and other complex physical systems. 
The idea is to utilize the natural tendency of the physical system to seek low-energy equilibria. 
This is precisely what underpins, for instance, quantum annealing \citep{bapst2013quantum}, coherent Ising machines \citep{meirzada2022lightsolver,wirzberger2023lightsolver,Yamamoto2017}, and even neural network-based approaches such as Hopfield networks \citep{Fahimi2021,Liang1996} and spiking networks \citep{lu2023combinatorial}. 

\begin{figure}[t!]
    \centering
    \includegraphics[width=\columnwidth]{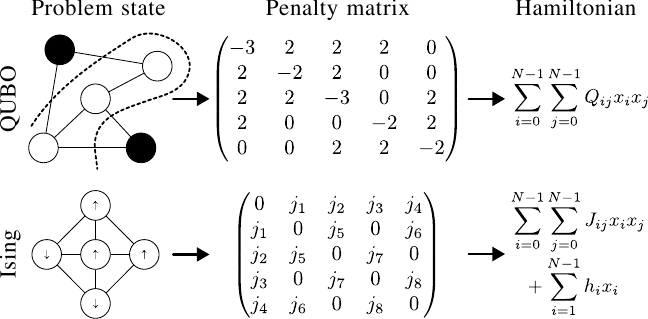}
    \caption{Schematic of the QUBO and Ising spin models.}
    \label{fig:qubo_ising_schema}
\end{figure}
In general, spin-glass models define an energy functional on a set of $N$ (spin) particles $\{x_i\}$, each of which can occupy one of a set of discrete states. The Hamiltonian (total energy) consists of self and interaction terms up to some order,
and induces a natural ``gradient flow'' toward lower energy states. In quantum settings, for example, the system famously evolves under Schrödinger's equation until it settles into an equilibrium configuration. In many combinatorial problems, the energy landscape has numerous local minima, and one often employs a homotopy strategy (i.e., a smoother Hamiltonian is gradually deformed into the more challenging target Hamiltonian) to improve the chances of finding the global minimum. 

A particularly popular spin-glass model relevant to combinatorial optimization is the \textit{Ising model}. It describes a lattice of magnetic dipoles/spins (for a $1$D chain of Ising spins, see \Cref{fig:spin}) that may take one of two discrete orientations. Its Hamiltonian can be written as:
\begin{equation}
\label{eq:ising}
    H_{\rm Ising} =  \sum_{i=0}^{N-1}\sum_{j=0}^{N-1} J_{ij} x_i x_j+ \sum_{i=1}^{N-1} h_ix_i,
\end{equation}
where $x_i \in \{\pm 1\}$ (or $\{0,1\}$, depending on conventions) are the spin variables and $J$ is a symmetric matrix of pairwise interaction coefficients.
The second summand in the Ising Hamiltonian is the ``magnetic field term'' (with $h$ being a vector of local fields) that can incorporate penalty or constraint enforcement if mapped appropriately from an optimization problem, such as the MaxCut (\cref{def:maxcut}). 

By encoding the cut as a spin arrangement, minimizing $H_{\rm Ising}$ aligns with finding an optimal partition. This has been the traditional way to approach CO problems with quantum computers, either with quantum annealing, as mentioned earlier, or with variational quantum algorithms such as the Quantum Approximate Optimization Algorithm (QAOA)~\citep{farhi2014quantum}. The connecting characteristic of all the above is that the problem is first formulated as a \textit{quadratic unconstrained binary optimization} (QUBO)~\citep[][]{glover2019quantum} problem which has a natural mapping to the Ising model \citep{Lucas2014}, as shown in \cref{fig:qubo_ising_schema}.
Such an approach has practical applications for industrial-scale problems~\citep{Buonaiuto2023, Giron2023, Matsumori2022} where classic methods, such as the B\&B, are not sufficient.



\subsection{Graph Neural Networks}
\label{sec:gnns}
The outlined close relationship between CO
problems and graph representations (\cref{sec:co}) highlights important characteristics that a solving approach should address.
First, the problem representation should be {\textit{permutation invariant}}, as the problem's structure---defined by the graph's connectivity---is essential, while the serialized order of its specification is irrelevant.
Second, combinatorial optimization typically involves graphs of varying sizes and shapes. Therefore, a representation invariant to such structural changes is highly desirable.

These invariance properties give rise to \textit{Graph neural networks} (GNNs)---a class of geometric deep learning~\citep{bronstein2021geometric} architectures designed to directly process data represented as graphs~\citep{scarselli2008graph}, which is typically more efficient than converting the problem into alternative forms.
One of the original, particularly efficient types of GNNs is the graph convolutional network (GCN)~\citep[][]{kipf2016semi}, which has previously been used for CO~\citep{schuetz2022combinatorial}.
To process the problem graphs, GCNs utilize the edges between nodes to form a ``message-passing'' scheme, where each node $v$ computes its representation $\mathbf{h}$ at layer $k$ based on aggregating ($\text{AGG} \in \{sum,\ max,\dots\}$) representations from its neighbors $u$ at layer $k-1$.
Formally, for a graph $ \mathcal{G} = (\mathcal{V}, \mathcal{E}) $, the representation $ \mathbf{h}_v^{(k)} $ of a node $ v \in \mathcal{V} $ at layer $ k $ is updated as: 
\begin{equation}
\mathbf{h}_v^{(k)} = \text{AGG}\left(\left\{\phi\left(\mathbf{h}_u^{(k-1)}
\right) \mid u \in \mathcal{N}(v)\right\}\right),
\end{equation}
where $ \mathcal{N}(v) $ denotes the neighbors of $ v $, and $\phi$ is a learnable transformation.
By applying the same transformation $\phi$ to all nodes, this scheme ensures that the models are permutation invariant to node ordering and are applicable to graphs of varying structures.

\begin{figure}[b]
    \centering
    \includegraphics[width=0.88\columnwidth]{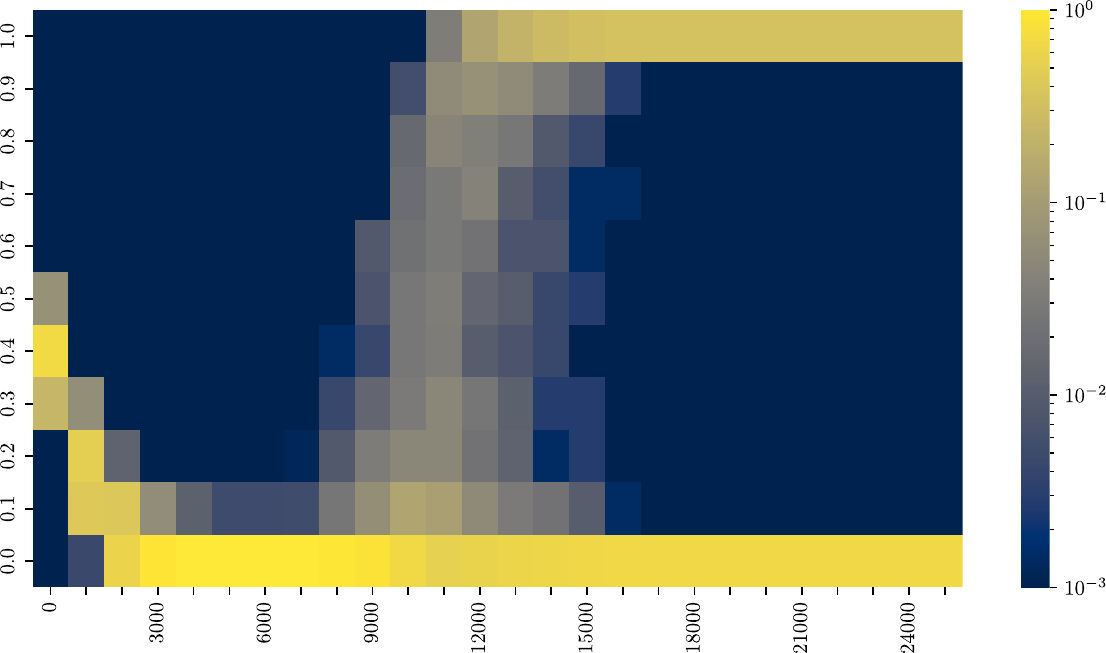}
    \caption{Distribution of post-activation values ($a^{post}$) over training epochs on the MaxCut problem with \textit{sparser} graphs.}
    \label{fig:maxcut_baseline_3_logits}
\end{figure}

\paragraph{GNN Approaches to CO}
\label{sec:taxonomy}

The intersection of deep learning and CO has recently been extensively studied. A review of~\citet{BENGIO2021405} outlines neural approaches relevant to solving combinatorial problems, including recurrent and attention-based architectures.
Focusing on GNNs, \citet{cappart2023combinatorial} then provided a detailed survey categorizing the GNN approaches into (i) {supervised}, (ii) {unsupervised}, and (ii) {reinforcement learning} based. Supervised methods learn binary assignments or heuristic values using labeled datasets. Unsupervised methods minimize loss functions encoding optimization objectives and constraints, bypassing the need for labels. Reinforcement learning employs GNNs to approximate value functions.

\paragraph{Unsupervised Training}
Solving CO problems with \textit{unsupervised} learning generally received the most attention.
\citet{karalias2020erdos} proposed a framework with probabilistic guarantees, later extended with principled objective relaxation~\citep{wang2022unsupervised} and meta-learning~\citep{wang2023unsupervised}.
Orthogonally, the idea of PI-GNNs~\citep{schuetz2022combinatorial, schuetz2022graph} was developed, followed by enhancements including reinforcement learning~\citep{rizvee2023graph}, automated machine learning~\citep{liu2024combinatorial}, recurrent updates~\citep{pugacheva2024enhancing}, quantum annealing~\citep{he2024quantum, loyola2024annealing}, efficient graph reduction~\citep{tran2025scalable}, and generalization to mixed-integer linear programming~\citep{ye2025designing}. Finally, \citet{ichikawa2024controlling} focused on annealing the continuous relaxation via a loss term.

\begin{figure}[b]
    \centering
    \includegraphics[width=0.97\columnwidth]{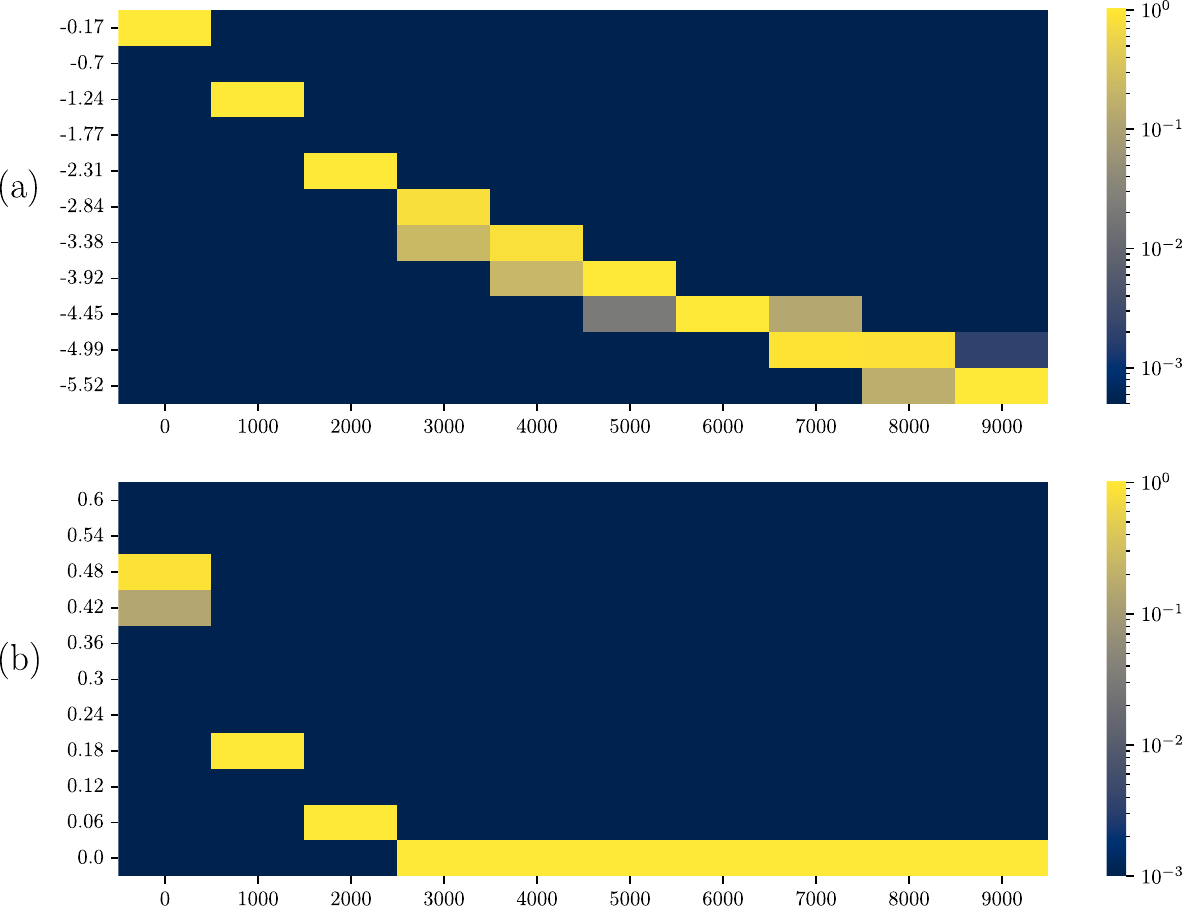}
    \caption{Distribution of (a) pre-activations ($a^{pre}$) and (b) post-activations ($a^{post}$) over training epochs on the MaxCut problem with \textit{denser} graphs.}
    \label{fig:maxcut_baseline_50_2x2}
\end{figure}

\begin{figure*}[t]
    \centering
    \includegraphics[width=\linewidth]{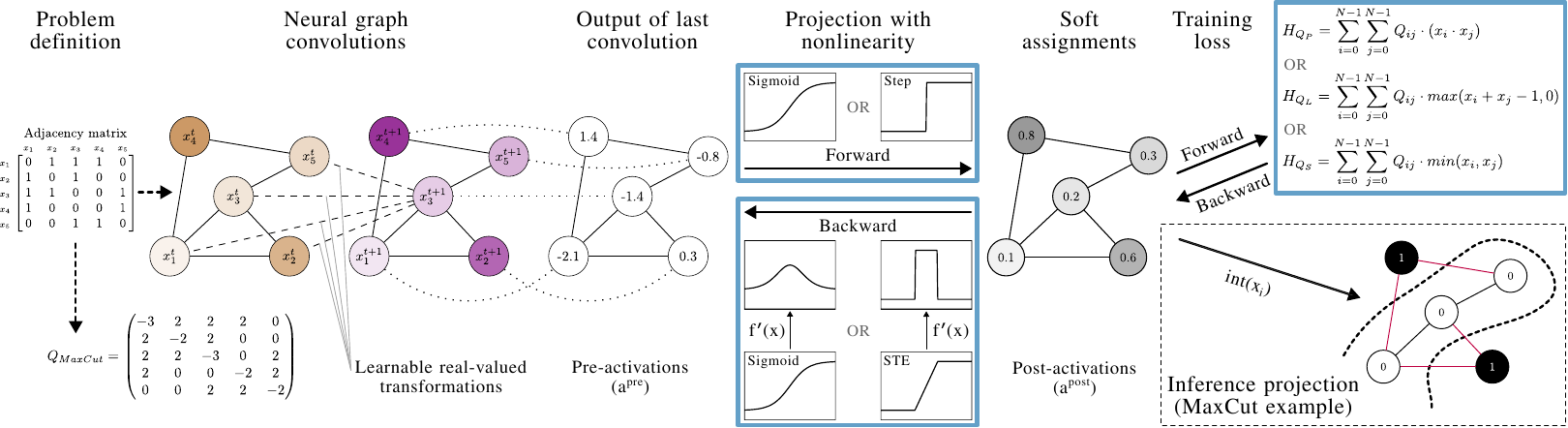}
    \caption{Schema of the PI-GNNs pipeline for solving CO under QUBO formalization (problem encoding, graph convolutions, node assignments), showcased on a MaxCut problem. The proposed modifications---binarization of the ultimate nonlinearity (\cref{sec:discretization}) and loss fuzzification (\cref{sec:fuzzy})---are highlighted in blue.}
    \label{fig:schema}
\end{figure*}

\section{The Approach}
\label{sec:methodology}
As outlined in~\cref{sec:introduction}, the performance of the PI-GNNs declines abruptly with the increasing density of the CO graphs,
diminishing the promise of scaling to the more difficult problems through GPU parallelization.
The contribution of this paper is to (i) provide an analysis of the issue (\cref{sec:motivation}) and (ii) propose improvements (\cref{sec:fuzzy,sec:discretization}) to advance to the denser CO settings.

As a problem encoding, we consider the {quadratic unconstrained binary optimization} (QUBO; \cref{sec:spin}) model~\citep[][]{glover2019quantum} which unifies a variety of \textit{NP}-hard CO problems, forming the basis for the 
PI-GNNs~\citep{schuetz2022combinatorial} framework we build upon in this work. The cost function for a QUBO problem can be expressed in compact form with the Hamiltonian:
\begin{equation}\label{eq:quboloss}
    H_{Q} = x^T Q x = \sum_{i=0}^{N-1}\sum_{j=0}^{N-1} Q_{ij}x_ix_j.
\end{equation}
Here, $x \in \{0,1 \}^N$ is the vector of binary decision variables, and the QUBO matrix $Q$ is a square matrix that encodes the objective of the original problem~\citep{glover2019quantum}.
Without loss of generality, the $Q$-matrix can be assumed to be symmetric or in upper triangular form. This Hamiltonian is famously related to the Ising spin model Hamiltonian (\cref{eq:ising}, \cref{sec:spin}).
Its encoding (\cref{eq:quboloss}) for an example MaxCut problem (\cref{def:maxcut}) is displayed in~\cref{fig:qubo_ising_schema}.

With the PI-GNNs, an instance of a combinatorial optimization problem
(\cref{def:co}) is then given by a graph $\mathcal{G} = (\mathcal{V}, \mathcal{E})$ with an 
associated adjacency matrix $A_\mathcal{G}$, and a cost function specified by the QUBO Hamiltonian $H_{Q}$.
Input features (embeddings) of nodes $\mathcal{V}$ in $\mathcal{G}$ are randomly initialized, and fed together with $A_\mathcal{G}$ to a standard GNN model (\cref{sec:gnns}) trained against the $H_{Q}$ loss minimization. 
During a single inference step, the graph is passed through multiple convolutional layers with intermediate ReLU nonlinearities.
After the final graph layer, the (scalar) node outputs ($a^{pre}$) are projected into the $(0, 1)$ interval using a logistic sigmoid $\sigma(a^{pre})$, serving as ``soft node assignments'' $a^{post}$ entering the loss $H_{Q}$.
Once the training converges, these final output values $x_i:=a_i^{post}$ are thresholded as ${\rm int}(x_i)$ to produce binary value assignments for the original CO problem.

\subsection{Analyzing PI-GNNs Training Dynamics}
\label{sec:motivation}

To analyze the declining performance of PI-GNNs with increasing graph density,
we tracked the model's intermediate and final outputs during the training.
Particularly, we tracked the ``pre-activation'' ($a^{pre}$) outputs of the final graph layer, and the ``post-activation'' ($a^{post}$) outputs of the final non-linearity, before the ultimate rounding (see~\cref{fig:schema}).
The evolution of activations ($a^{post}$) during training for one of the \textit{sparser} settings is visualized as a heatmap, where color-encoded histograms represent the distribution of activation values over training epochs,\footnote{Frequencies are computed from training runs on $20$ graphs.} shown in~\cref{fig:maxcut_baseline_3_logits} and in~\cref{app:dynamics}.
The analysis indicates that, starting from random initial assignments, nodes (variables) first undergo a \textit{nullification} phase during the initial thousands of iterations, minimizing the positive terms of the QUBO loss associated with constraint violations. Subsequently, a subset of nodes gradually becomes active in a later phase, increasing the negative terms of the QUBO loss that contribute to the solution value.

The later stage is, however, absent in the training dynamics on \textit{denser} graphs ($d \in \{20, 30, 40, 50\}$), resulting in a degenerated all-zero assignment, as shown in~\cref{fig:maxcut_baseline_50_2x2} (b).\footnote{The difference in epoch range (cf.~\cref{fig:maxcut_baseline_3_logits}) is caused by early stopping.}
To inspect this failure mode more closely, a heatmap of pre-activation values from the same experiment is plotted in~\cref{fig:maxcut_baseline_50_2x2} (a).

Stagnation in training dynamics can typically be attributed to the presence of multimodality, that is, multiple wells of suboptimal local minima, as well as locally flat ``saddle'' neighborhoods. The greater the multimodality, the greater the chance of a suboptimal yet sufficiently steep well-trapping gradient flow. 
There is a natural correspondence to this observation in the literature on modeling combinatorial optimization as spin-glass energy (\cref{sec:spin}). \citet{cheeseman1991really} observed that the vast majority of hard complexity combinatorial problems are, in practice, not especially problematic to solve. However, one can observe that certain perturbations of the combinatorial problem definition lead to starkly more difficult problems to solve, whose actual computational load for approaching a solution begins to approach the worst-case complexity. 
Simultaneously, along this perturbation, the corresponding energy of the spin glass model describing the combinatorial problem undergoes phase transitions, complicating the topology of the critical points of the objective landscape. 
We hypothesize that increasing graph connectivity, and thus the variable interdependencies, constitutes such a perturbation.


\subsection{Fuzzy Logic Interpretation of the QUBO Relaxation}
\label{sec:fuzzy}

As outlined in~\cref{fig:qubo_ising_schema}, the cost function from~\cref{eq:quboloss} directly encodes graph optimization problems with binary-valued variables $x_i$.
The interpretation is that the off-diagonal elements of the $Q$ matrix ($Q_{ij}, i \neq j$) represent the penalties/rewards for \textit{edges} included in the solution set, i.e., the penalty is added if both nodes are selected: $(x_i = 1) \land (x_j = 1)$. We will further use a simplified notation $x_i \land x_j$ or its algebraic version $x_i \cdot x_j$. The on-diagonal elements represent the penalties/rewards for \textit{nodes} from the solution set, and can also be selected by a (trivial) $x_i \land x_i$.

In the relaxed PI-GNN version, the variables become \textit{real-valued} $x_i \in [0, 1]$, corresponding to ``soft'' assignments (node selections). To address the aforementioned problematic phase shift from $0$ to $1$ (\cref{sec:motivation}), we propose to inspect this binary relaxation in a more principled manner. Specifically, we propose to model the assignments as \textit{fuzzy values} representing \textit{membership degrees} $\mu(x_i)$ of nodes to the solution set (\cref{def:co}).
With that interpretation, each term for the elements of matrix $Q$ now expresses a \textit{fuzzy conjunction}.
\begin{definition}[Fuzzy Conjunction]
Let $x_i, x_j \in [0,1]$ be fuzzy truth values. A \emph{fuzzy conjunction} $x_i \underset{\cdot}{\wedge} x_j$ is a generalization of logical conjunction modeled by a \emph{$t$-norm} $T: [0,1] \times [0,1] \to [0,1]$ that satisfies the following properties: (i) commutativity, (ii) associativity, (iii) monotonicity, and (iv) boundary conditions.
\end{definition}



\citet{schuetz2022combinatorial} intuitively defaulted to the algebraic version of the classical conjunction $x_i \cdot x_j$, corresponding to a product fuzzy conjunction $x_i \landp x_j$. However, there is a number of other common $t$-norms, such as the standard $x_i \lands x_j = \min \{x_i, x_j\}$ or Łukasiewicz conjunction $x_i \landl x_j = \max \{x_i + x_j - 1, 0\}$, which can be more appropriate in different scenarios, including QUBO. The generalized fuzzy form of the Hamiltonian then reads:

\begin{equation}\label{eq:qubofuzzy}
    H_{Q_F} = \sum_{i=0}^{N-1}\sum_{j=0}^{N-1} Q_{ij} \cdot (x_i \underset{\cdot}{\wedge} x_j).
\end{equation}

This defines a QUBO loss that evaluates to the exact same values as the original form from~\citet{schuetz2022combinatorial} for binary-valued decisions,
while presenting distinct landscapes for different fuzzy relaxations.
Compare now, analytically, the form of the two losses for the product and Łukasiewicz fuzzy conjunction: 
\begin{equation}
 Q_{ij} \cdot (x_i \cdot x_j) \text{~~~compared to~~~} \\  Q_{ij} \cdot \max\{x_i+x_j-1,  0\} 
\end{equation}
The landscape contribution for each term in the first Hamiltonian will be a convex-bowed increasing function as $x_i$ and $x_j$ increase from $0$ to $1$. For the second Hamiltonian, the loss is piecewise linear. 

Recall the seminal work on training feedforward neural networks by modeling loss as a spin glass~\citep{glorot2010understanding}. This work established that the training objective landscape had an exponentially increasing number of suboptimal saddle points that slowed training dynamics. One can see that as the number of nonzero $Q_{ij}$, i.e., the connectivity, increases, the number of bilinear terms in the product formulations increases, that is, the larger quantity of off-diagonal terms in $Q_{ij}$. As a result, the inertia, that is, the number of positive and negative eigenvalues of $Q$, becomes more variable across the relaxed feasible region $[0,1]^N$ and, similarly to standard feedforward neural networks, the number of saddle points increases. At the same time, since $N$ remains constant, the network becomes less and less overparametrized. Thus, the chances of stochastic gradient dynamics stagnating at saddle points and the presence of suboptimal local minimizers both increase, presenting a significantly more challenging problem to solve.

Consider, by contrast, the fuzzy Lukasiewicz conjunction. This has two contrasting features. First, it is piecewise linear, and thus has no curvature, obviating the potential of producing saddle points. Second, for $x_i+x_j\le 1$, which amounts to half of the measure of $[0,1]\times[0,1]$, the objective component is entirely flat at zero. This creates a structural sparsity in the dependence of the objective with respect to the inputs. Both of these factors mitigate some of the landscape complexity introduced by an increasing number of nonzero $\{Q_{ij}\}$~\citep{davis2022escaping}. One can also observe that this activation resembles the topography of the piece-wise linear Rectified Linear Units. After~\citep{glorot2010understanding}, the use of these activations became more favored in NN models, observed to exhibit better performance and more stable training.
See further~\citet{gombotz2021fuzzy,mueller1988fuzzy,shafee2002spin} for applications of fuzzy conjunctions in spin glass models.

\subsection{Binarizing PI-GNNs}
\label{sec:discretization}

A related alternative to the fuzzy relaxation of the QUBO Hamiltonian is to avoid the PI-GNN's soft assignment step in the first place by forcing the model to directly produce binary values.
Although neural networks are inherently continuous, since their training relies on methods that assume continuity and the existence of gradients, there are approaches that enable them to exhibit discrete behaviors.

\paragraph{Binarized Neural Networks}
\textit{Binarized neural networks} (BNNs)~\citep[][]{hubara2016binarized} are a class of deep learning models with weights and activations constrained to binary values. Originally, BNNs have been constrained to values $1$ and $-1$ but, similarly to the QUBO and Ising models (\cref{sec:spin}), the methods can be adjusted to other pairs of values $(a, b)$, such as the logical $(0, 1)$. The primary contribution of BNNs is that the transformations over purely binary matrices (implementable by bitwise operations) can reduce memory consumption and computational requirements~\citep{simons2019review}. However, importantly for this work, the binarized outputs also naturally represent discrete problems, avoiding the need for the relaxation to real values.
Exploiting the latter advantage, we aim to binarize the output PI-GNNs activations, particularly the logistic sigmoid function $\sigma(x) = \frac{1}{1+e^{-x}}$ used by~\citet{schuetz2022combinatorial}, resulting in a step function:
\begin{equation}
    \sigma_{\rm step}(x) = \begin{cases}
        a, & \text{if $x \leq 0$,}\\
        b, & \text{otherwise.}
    \end{cases}
\end{equation}
However, the step function is discontinuous, and its gradient, where defined, is zero. Thus, for the \textit{backward pass}, a continuous approximation with a non-zero gradient is required. A practical solution, which we adopt from BNNs, is the ``straight-through estimator'' (STE)~\citep[][]{bengio2013estimating}:
\begin{equation}
\label{eq:stepste}
    \sigma_{\rm STE}(x) = {\rm clip}(x, a, b) = \max(\min(x, b), a).
\end{equation}

\begin{table*}[t]
\caption{Table of results for the maximum cut problem with $N=1000$ nodes, w.r.t. the best-of-N metric}
\label{table:maxcut-result}
\resizebox{\textwidth}{!}{%
\begin{tabular}{ccccccccccccccc}
\toprule
 \multirow{2}{*}{MaxCut BoN} & \multicolumn{2}{c}{$d=3$} & \multicolumn{2}{c}{$d=5$} & \multicolumn{2}{c}{$d=10$} & \multicolumn{2}{c}{$d=20$} & \multicolumn{2}{c}{$d=30$} & \multicolumn{2}{c}{$d=40$} & \multicolumn{2}{c}{$d=50$} \\\cmidrule(lr){2-3}\cmidrule(lr){4-5}\cmidrule(lr){6-7}\cmidrule(lr){8-9}\cmidrule(lr){10-11}\cmidrule(lr){12-13}\cmidrule(lr){14-15}
& BoN     & RR\textsubscript{BoN} & BoN     & RR\textsubscript{BoN} & BoN     & RR\textsubscript{BoN} & BoN     & RR\textsubscript{BoN} & BoN     & RR\textsubscript{BoN} & BoN     & RR\textsubscript{BoN} & BoN     & RR\textsubscript{BoN} \\\midrule
\texttt{baseline} & 1273.90 & 0.36 & 1861.70 & 0.21 & 3144.10 & 0.33 & 0.00 & 0.25 & 0.00 & 0.25 & 0.00 & 0.25 & 0.00 & 0.25 \\
\texttt{temp-lin} & 1277.35 & 0.56 & 1878.60 & 0.38 & 3165.00 & 0.43 & 0.00 & 0.25 & 0.00 & 0.25 & 0.00 & 0.25 & 0.00 & 0.25 \\
\texttt{temp-log} & \textbf{1283.10} & \textbf{0.86} & 1887.70 & 0.46 & 156.20 & 0.13 & 0.00 & 0.25 & 0.00 & 0.25 & 0.00 & 0.25 & 0.00 & 0.25 \\
\texttt{temp-exp} & 1274.25 & 0.36 & 1866.60 & 0.25 & 3153.90 & 0.34 & 0.00 & 0.25 & 0.00 & 0.25 & 0.00 & 0.25 & 0.00 & 0.25 \\
\texttt{bin-ste} & 1062.30 & 0.12 & 1469.05 & 0.12 & 2089.10 & 0.16 & 3987.40 & 0.65 & \textbf{5914.30} & \textbf{0.93} & 7719.40 & \textbf{0.95} & \textbf{9699.70} & \textbf{0.98} \\
\texttt{bin-sig} & 1180.65 & 0.15 & 1577.30 & 0.14 & 2064.20 & 0.16 & 3986.70 & 0.64 & 5911.20 & 0.92 & \textbf{7721.90} & 0.93 & 9699.00 & 0.98 \\
\texttt{fuzzy-std} & 1236.75 & 0.20 & \textbf{1932.90} & \textbf{1.00} & \textbf{3260.90} & \textbf{1.00} & 0.00 & 0.25 & 0.00 & 0.25 & 0.00 & 0.25 & 0.00 & 0.25 \\
\texttt{fuzzy-luk} & 1220.95 & 0.17 & 1698.15 & 0.17 & 2623.30 & 0.20 & \textbf{4122.70} & \textbf{0.80} & 5551.70 & 0.40 & 6967.20 & 0.37 & 8482.10 & 0.33 \\\bottomrule
\end{tabular}%
}
\end{table*}
\begin{table*}[h!]
\centering
\caption{Table of results for the maximum independent set problem with $N=1000$ nodes, w.r.t. the best-of-N metric}
\label{table:mis-result}
\resizebox{\textwidth}{!}{%
\begin{tabular}{ccccccccccccccc}
\toprule
 \multirow{2}{*}{MIS BoN} & \multicolumn{2}{c}{$d=3$} & \multicolumn{2}{c}{$d=5$} & \multicolumn{2}{c}{$d=10$} & \multicolumn{2}{c}{$d=20$} & \multicolumn{2}{c}{$d=30$} & \multicolumn{2}{c}{$d=40$} & \multicolumn{2}{c}{$d=50$} \\\cmidrule(lr){2-3}\cmidrule(lr){4-5}\cmidrule(lr){6-7}\cmidrule(lr){8-9}\cmidrule(lr){10-11}\cmidrule(lr){12-13}\cmidrule(lr){14-15}
& BoN     & RR\textsubscript{BoN} & BoN     & RR\textsubscript{BoN} & BoN     & RR\textsubscript{BoN} & BoN     & RR\textsubscript{BoN} & BoN     & RR\textsubscript{BoN} & BoN     & RR\textsubscript{BoN} & BoN     & RR\textsubscript{BoN} \\\midrule
\texttt{baseline} & 421.75 & 0.39 & 345.90 & 0.39 & 0.00 & 0.25 & 0.00 & 0.40 & 0.00 & 0.33 & 0.00 & 0.31 & 0.00 & 0.26 \\
\texttt{temp-lin} & 423.45 & 0.55 & \textbf{349.65} & \textbf{0.77} & 0.00 & 0.25 & 0.00 & 0.40 & 0.00 & 0.33 & 0.00 & 0.31 & 0.00 & 0.26 \\
\texttt{temp-log} & \textbf{426.85} & \textbf{0.89} & 348.75 & 0.62 & 0.00 & 0.25 & 0.00 & 0.40 & 0.00 & 0.33 & 0.00 & 0.31 & 0.00 & 0.26 \\
\texttt{temp-exp} & 422.25 & 0.45 & 347.15 & 0.44 & 0.00 & 0.25 & 0.00 & 0.40 & 0.00 & 0.33 & 0.00 & 0.31 & 0.00 & 0.26 \\
\texttt{bin-ste} & 227.25 & 0.03 & 237.93 & 0.11 & 17.10 & 0.50 & 8.00 & 0.50 & 10.15 & 0.62 & 8.50 & 0.62 & 8.00 & 0.70 \\
\texttt{bin-sig} & 347.94 & 0.15 & 38.60 & 0.03 & 11.10 & 0.49 & 8.00 & 0.50 & 10.15 & 0.62 & 8.50 & 0.62 & 8.05 & 0.73 \\
\texttt{fuzzy-std} & 413.30 & 0.20 & 336.85 & 0.20 & 0.00 & 0.25 & 0.00 & 0.40 & 0.00 & 0.33 & 0.00 & 0.31 & 0.00 & 0.26 \\
\texttt{fuzzy-luk} & 394.50 & 0.02 & 303.86 & 0.12 & \textbf{189.40} & \textbf{1.00} & \textbf{84.50} & \textbf{1.00} & \textbf{23.10} & \textbf{0.90} & \textbf{17.45} & \textbf{0.87} & \textbf{11.15} & \textbf{0.80} \\\bottomrule
\end{tabular}%
}
\end{table*}

\begin{figure*}[ht!]
    \centering
    \includegraphics[width=0.85\linewidth]{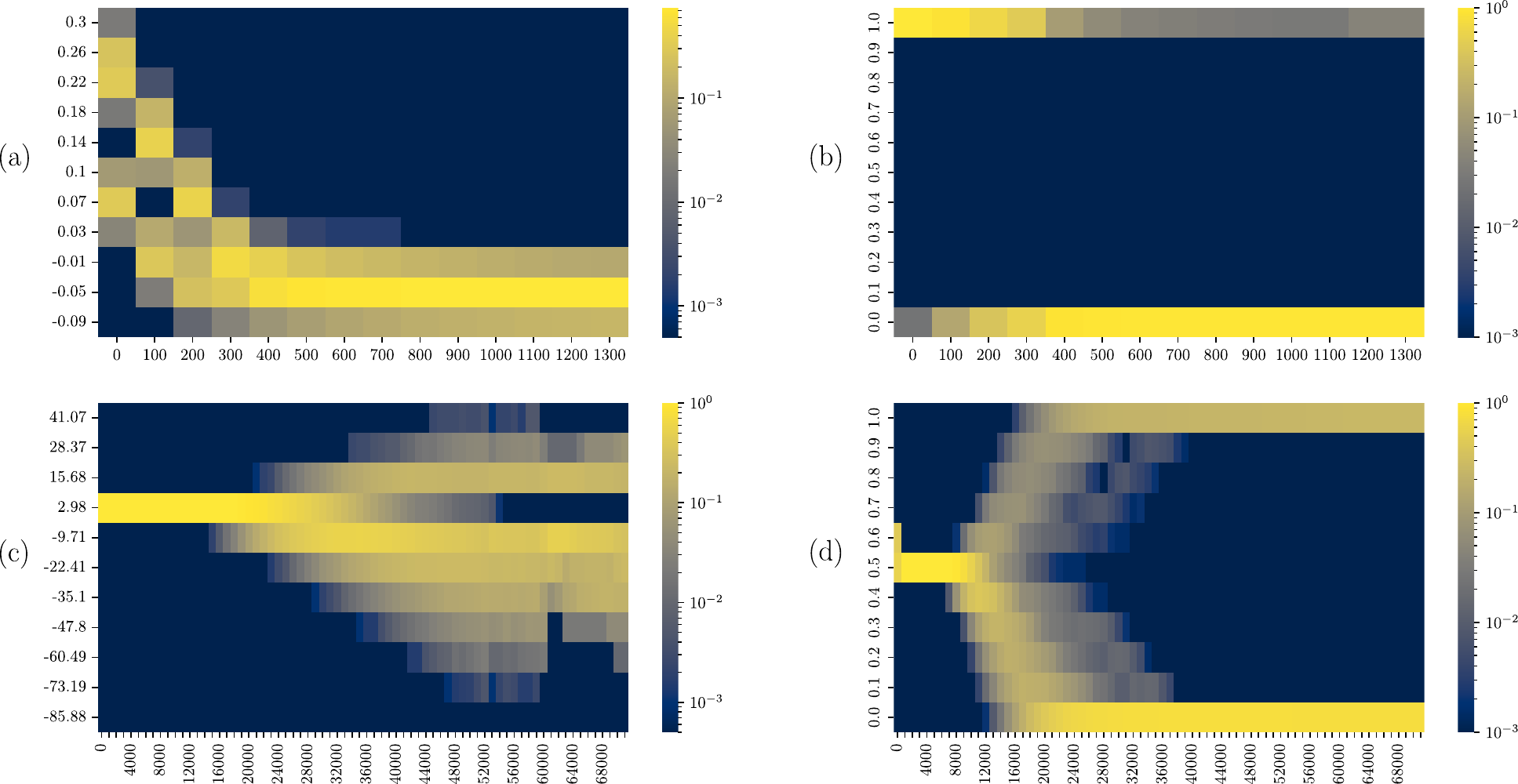}
    \caption{Distributions of GNN's pre-activation ($a^{pre}$) and post-activation ($a^{post}$) values over training epochs on $50$-regular MaxCut graphs: (a) $a^{pre}$ for \texttt{bin-ste}, (b) $a^{post}$ for \texttt{bin-ste}, (c) $a^{pre}$ for \texttt{fuzzy-luk}, (d) $a^{post}$ for \texttt{fuzzy-luk}.}
    \label{fig:values-2x2}
\end{figure*}

Alternatively, the original sigmoid function $\sigma(x)$ can also be used in the {backward} pass. 
In summary, we propose the following forward and backward pass modifications to the PI-GNN models for CO:
\begin{equation}
\label{eq:forback}
\begin{gathered}
    \text{Forward: }\ f(x) = \sigma_{\rm step}(x)\\
    \text{Backward:}\ f'(x) = \sigma_{\rm STE}'(x)\ \ \text{~~or~~}\ \ \sigma'(x)
\end{gathered}
\end{equation}

\paragraph{Temperature Annealing}
An alternative approach for discretization of activation functions, avoiding misalignment between forward and backward passes, is ``soft discretization'' via \textit{temperature annealing}~\citep{bonnell2011implementation, gold1996softmax, papernot2021tempered}. 
In this method, the slope of the sigmoid function gradually increases as the training progresses, approaching the step function in the limit: $ \lim_{i \to \infty} \sigma_i(x) = \sigma_{\text{step}}(x) $. A high slope would normally cause vanishing gradient issues, but with the adjusted sigmoid
\begin{equation}
    \sigma_i(x) = \frac{1}{1+e^{-\frac{x}{T_i}}}
\end{equation}
the temperature $T_i$ is gradually decreased (annealed), allowing the function to converge to discrete outputs while maintaining sufficient gradients in the early training stages.
Practically, it is easier to implement a sigmoid with an inverse temperature and gradually increase its value, starting at $\frac{1}{T_0} = 1$ and ending at $\frac{1}{T_n} = \#epochs$.
Temperature annealing further requires choosing a \textit{schedule} determining the pace of the temperature change. We consider three types of schedules w.r.t. the inverse temperature: (i) \textit{linear} $\frac{1}{T_i} = i$, (ii)~\textit{logarithmic} $\frac{1}{T_i} = log_2(i)$, and (iii) \textit{exponential}  $\frac{1}{T_i} = 2^i$.

\section{Experiments}
\label{sec:experiments} 
The experiments
were aligned with the original methodology of PI-GNNs~\citep{schuetz2022combinatorial} by choosing equivalent CO problems, data collection approach, and overall research setup.
The purpose of the experiments is to address the problematic PI-GNN training performance for denser CO 
settings (\cref{sec:motivation}) with the methods introduced in~\cref{sec:fuzzy,sec:discretization}.

\paragraph{Methods}
Particularly, apart from the original PI-GNN approach (\texttt{baseline}), we propose a portfolio of the following $7$ methods based on the concepts of (i) soft (annealing) and (ii) hard binarization (\cref{sec:discretization}), and (iii) fuzzy relaxation (\cref{sec:fuzzy}):
\begin{enumerate}
    \item adjusting sigmoid with linearly scheduled inverse temperature (\texttt{temp-lin})
    \item adjusting sigmoid with logarithmically scheduled inverse temperature (\texttt{temp-log})
    \item adjusting sigmoid with exponentially scheduled inverse temperature (\texttt{temp-exp})
    \item binarizing sigmoid in the forward pass, using straight-through estimator in the backward pass (\texttt{bin-ste})
    \item binarizing sigmoid in the forward pass, using original sigmoid in the backward pass (\texttt{bin-sig})
    \item changing fuzzy conjunction in the QUBO loss to the standard conjunction (\texttt{fuzzy-std})
    \item changing fuzzy conjunction in the QUBO loss to the Łukasiewicz conjunction (\texttt{fuzzy-luk}).
\end{enumerate}

\citet{schuetz2022combinatorial} argued that $N$ experiments with different random seeds can be run in parallel and the best run can be selected as the solution.\footnote{Unlike in supervised learning, we can order the quality of predictions without knowing the ground-truth by comparing the value/cost of feasible solutions.}
We directly extend upon this principle by running a \textit{portfolio} of the above proposed models in the experiments.

\paragraph{Training}
For a direct comparison with~\citet{schuetz2022combinatorial}, we make use of the same GNN ansatz, with each training experiment being repeated over $N=5$ random initializations. 
The GNN model is iteratively trained on individual graph instances against a custom loss function that encodes a relaxed version of the optimization problem, as specified by the Hamiltonian $H_Q$ (\cref{fig:schema}).
The optimization hyperparameters are maintained from~\citet{schuetz2022combinatorial}, using the Adam optimizer~\citep{kingma2014adam} and early stopping in case of no improvement.


\paragraph{Data}
Aiming at improving the performance of PI-GNNs over graphs of increasing connectivity, we
vary the density of the graphs $\mathcal{G}$, controlled by the regularity parameter $d \in \{3, 5, 10, 20, 30, 40, 50\}$.
Otherwise, the data collection routine remains identical to the one introduced by~\citet{schuetz2022combinatorial}, generating $20$ random graphs for each graph type (combination of problem and $d$-regularity).

\paragraph{Metrics}
We define the following metrics to evaluate the proposed architectures.
Firstly, the $N$ random seeds can be aggregated by choosing the best-of-N experiment ($BoN = \max_{i=1..N} {\rm GNN}(\theta, \mathcal{G})$), or by averaging the values of the runs ($Avg = \frac{1}{N} \sum_{i=1..N} {\rm GNN}(\theta, \mathcal{G})$).
Further, the two metrics retrieving aggregated cost of solutions are also accompanied by their mean reciprocal ranking ($MRR = \frac{1}{Q} \sum_{i=1..Q} \frac{1}{{\rm rank}_i}$ for $Q:= \#architectures$).

\subsection{Results}
The portfolio of the proposed methods, using different discretization and fuzzification techniques, outperformed the baseline QUBO on all datasets, measured with both best-of-N ($BoN$) and average ($Avg$) metrics. 
See~\cref{table:maxcut-result} and~\cref{table:mis-result} for $BoN$ in the MaxCut and MIS domains.
Specifically, the soft discretization techniques based on temperature annealing (\texttt{temp-lin}, \texttt{temp-log}, \texttt{temp-exp}) slightly outperformed the \texttt{baseline} on sparse graphs, while the binarization and fuzzy-logic adjustments clearly dominated all the denser setups.
Similar results then prevail when upscaling the graphs.

\paragraph{Training dynamics}
Many of the proposed architectures also exhibited improved training dynamics (\cref{sec:motivation}).
Specifically, the pre-activations of binarized sigmoids (\texttt{bin-ste} and \texttt{bin-sig}) remained appropriately in the \textit{proximity of zero}, and the Łukasiewicz conjunction also converged more smoothly in most cases compared to the baseline. 
The pre- and post-activation values for \texttt{bin-ste} and \texttt{fuzzy-luk} are depicted in~\cref{fig:values-2x2}.

For a complete overview of training hyperparameters, scaling results with varying graph sizes $n \in \{100, 1000, 10000\}$, and evaluation with the $Avg$ metric, see~\cref{app:misresults,app:scaling,app:resavg,app:hyperparams}. The source code is available at~\url{https://github.com/martin-krutsky/gnn-comb-opt}.

\subsection{Discussion}

Analyzing the results, we confirm that the collapse of the original PI-GNNs on dense graphs can be attributed to (i) the increased ratio between positive and negative terms of QUBO pressuring the model towards more zero predictions, and (ii) the propensity of sigmoid networks to confident predictions far from the classification boundary, reducing their ability for prediction shifts.
While the first issue is difficult to overcome without losing the optimality of the QUBO framework, our methods contribute to overcoming the latter.
The binarized versions of sigmoid allow minimizing the positive terms of QUBO loss without leaving the active regions of the input space near the decision boundary.
Similarly, the Łukasiewicz conjunction does not push the predictions far from the decision boundary since, once the conjunction reaches the zero plateau, the loss remains minimal.

\section{Conclusion}
\label{sec:conclusion}

We introduced improvements to PI-GNNs for combinatorial optimization, leveraging fuzzy interpretation of the QUBO Hamiltonian and concepts from binarized neural networks.
Our analysis uncovered that a phase transition in the model's activation distribution is responsible for the degrading performance of the original PI-GNNs on denser problems.
We argued for discretization techniques and fuzzy relaxations of QUBO to resolve the issue, and proposed respective adjustments to the PI-GNN algorithm.
The portfolio of adjusted architectures was comprehensively tested on datasets of increasing graph connectivity, and its superior performance over the baseline was confirmed, especially in denser settings.
This indicates that more principled deep learning discretization approaches can be a promising building block for models dealing with relaxed CO problems.

\section*{Ethics Statement}

This paper presents an improvement of a deep learning method for solving combinatorial 
optimization problems. While our work may have broad societal implications through its applications in industrial automation, such impacts are already widely recognized within the combinatorial optimization community. We therefore believe no specific ethical concerns require emphasis in this context.



\begin{ack}
This project was supported by the Czech Science Foundation grant No. 24-11664S.
GK also acknowledges support by project MIS 5154714 of the National Recovery and Resilience Plan Greece 2.0 funded by the European Union under the NextGenerationEU Program.
\end{ack}

\section*{Disclaimer}
This paper was prepared for information purposes and is not a product of HSBC Bank Plc. or its affiliates. Neither
HSBC Bank Plc. nor any of its affiliates make any explicit or implied representation or warranty and none of them accept any liability in connection with this paper, including, but not limited to, the completeness, accuracy, reliability of information contained herein and the potential legal, compliance, tax or accounting effects thereof. 



\bibliography{mybibfile}


\newpage
\appendix
\onecolumn

\section{MIS Results}
\label{app:misresults}
As the main text, due to space constraints, concentrated on presenting the results on the maximum cut problem, we report the results on the maximum independent set in~\cref{table:3,table:4}.
Similarly to MaxCut, the proposed discretization and fuzzy-logical adjustments outperform the \texttt{baseline}. 

\begin{table*}[h!]
\centering
\caption{Table of results for the maximum independent set problem with $N=100$ nodes, w.r.t. the best-of-N metric}
\label{table:3}
\resizebox{\textwidth}{!}{%
\begin{tabular}{ccccccccccccccc}
\toprule
 \multirow{2}{*}{MIS BoN} & \multicolumn{2}{c}{$d=3$} & \multicolumn{2}{c}{$d=5$} & \multicolumn{2}{c}{$d=10$} & \multicolumn{2}{c}{$d=20$} & \multicolumn{2}{c}{$d=30$} & \multicolumn{2}{c}{$d=40$} & \multicolumn{2}{c}{$d=50$} \\\cmidrule(lr){2-3}\cmidrule(lr){4-5}\cmidrule(lr){6-7}\cmidrule(lr){8-9}\cmidrule(lr){10-11}\cmidrule(lr){12-13}\cmidrule(lr){14-15}
& BoN     & RR\textsubscript{BoN} & BoN     & RR\textsubscript{BoN} & BoN     & RR\textsubscript{BoN} & BoN     & RR\textsubscript{BoN} & BoN     & RR\textsubscript{BoN} & BoN     & RR\textsubscript{BoN} & BoN     & RR\textsubscript{BoN} \\\midrule
\texttt{baseline} & 42.70 & 0.71 & 29.05 & 0.62 & 0.00 & 0.25 & 0.00 & 0.25 & 0.00 & 0.26 & 0.00 & 0.25 & 0.00 & 0.25 \\
\texttt{temp-lin} & \textbf{42.85} & \textbf{0.83} & 22.60 & 0.50 & 0.00 & 0.25 & 0.00 & 0.25 & 0.00 & 0.26 & 0.00 & 0.25 & 0.00 & 0.25 \\
\texttt{temp-log} & \textbf{42.85} & 0.79 & 2.20 & 0.15 & 0.00 & 0.25 & 0.00 & 0.25 & 0.00 & 0.26 & 0.00 & 0.25 & 0.00 & 0.25 \\
\texttt{temp-exp} & 42.75 & 0.75 & 29.25 & 0.64 & 0.00 & 0.25 & 0.00 & 0.25 & 0.00 & 0.26 & 0.00 & 0.25 & 0.00 & 0.25 \\
\texttt{bin-ste} & 26.55 & 0.14 & 22.55 & 0.19 & 6.75 & 0.53 & 4.90 & \textbf{0.85} & \textbf{2.85} & \textbf{0.83} & \textbf{3.15} & \textbf{0.90} & 2.35 & 0.78 \\
\texttt{bin-sig} & 27.60 & 0.13 & 11.15 & 0.17 & 6.70 & 0.52 & 4.90 & \textbf{0.85} & \textbf{2.85} & \textbf{0.83} & \textbf{3.15} & \textbf{0.90} & 2.35 & 0.78 \\
\texttt{fuzzy-std} & 42.05 & 0.47 & \textbf{35.05} & \textbf{0.86} & 0.00 & 0.25 & 0.00 & 0.25 & 0.00 & 0.26 & 0.00 & 0.25 & 0.00 & 0.25 \\
\texttt{fuzzy-luk} & 40.75 & 0.27 & 32.60 & 0.38 & \textbf{16.75} & \textbf{0.97} & \textbf{5.20} & 0.63 & 2.60 & 0.73 & 2.55 & 0.67 & \textbf{2.60} & \textbf{0.87} \\\bottomrule
\end{tabular}%
}
\end{table*}
\begin{table*}[h!]
\centering
\caption{Table of results for the maximum independent set problem with $N=1000$ nodes, w.r.t. the best-of-N metric}
\label{table:4}
\resizebox{\textwidth}{!}{%
\begin{tabular}{ccccccccccccccc}
\toprule
 \multirow{2}{*}{MIS BoN} & \multicolumn{2}{c}{$d=3$} & \multicolumn{2}{c}{$d=5$} & \multicolumn{2}{c}{$d=10$} & \multicolumn{2}{c}{$d=20$} & \multicolumn{2}{c}{$d=30$} & \multicolumn{2}{c}{$d=40$} & \multicolumn{2}{c}{$d=50$} \\\cmidrule(lr){2-3}\cmidrule(lr){4-5}\cmidrule(lr){6-7}\cmidrule(lr){8-9}\cmidrule(lr){10-11}\cmidrule(lr){12-13}\cmidrule(lr){14-15}
& BoN     & RR\textsubscript{BoN} & BoN     & RR\textsubscript{BoN} & BoN     & RR\textsubscript{BoN} & BoN     & RR\textsubscript{BoN} & BoN     & RR\textsubscript{BoN} & BoN     & RR\textsubscript{BoN} & BoN     & RR\textsubscript{BoN} \\\midrule
\texttt{baseline} & 421.75 & 0.39 & 345.90 & 0.39 & 0.00 & 0.25 & 0.00 & 0.40 & 0.00 & 0.33 & 0.00 & 0.31 & 0.00 & 0.26 \\
\texttt{temp-lin} & 423.45 & 0.55 & \textbf{349.65} & \textbf{0.77} & 0.00 & 0.25 & 0.00 & 0.40 & 0.00 & 0.33 & 0.00 & 0.31 & 0.00 & 0.26 \\
\texttt{temp-log} & \textbf{426.85} & \textbf{0.89} & 348.75 & 0.62 & 0.00 & 0.25 & 0.00 & 0.40 & 0.00 & 0.33 & 0.00 & 0.31 & 0.00 & 0.26 \\
\texttt{temp-exp} & 422.25 & 0.45 & 347.15 & 0.44 & 0.00 & 0.25 & 0.00 & 0.40 & 0.00 & 0.33 & 0.00 & 0.31 & 0.00 & 0.26 \\
\texttt{bin-ste} & 227.25 & 0.03 & 237.93 & 0.11 & 17.10 & 0.50 & 8.00 & 0.50 & 10.15 & 0.62 & 8.50 & 0.62 & 8.00 & 0.70 \\
\texttt{bin-sig} & 347.94 & 0.15 & 38.60 & 0.03 & 11.10 & 0.49 & 8.00 & 0.50 & 10.15 & 0.62 & 8.50 & 0.62 & 8.05 & 0.73 \\
\texttt{fuzzy-std} & 413.30 & 0.20 & 336.85 & 0.20 & 0.00 & 0.25 & 0.00 & 0.40 & 0.00 & 0.33 & 0.00 & 0.31 & 0.00 & 0.26 \\
\texttt{fuzzy-luk} & 394.50 & 0.02 & 303.86 & 0.12 & \textbf{189.40} & \textbf{1.00} & \textbf{84.50} & \textbf{1.00} & \textbf{23.10} & \textbf{0.90} & \textbf{17.45} & \textbf{0.87} & \textbf{11.15} & \textbf{0.80} \\\bottomrule
\end{tabular}%
}
\end{table*}

\section{Scaling the Experiments}
\label{app:scaling}
To compare the behavior of the proposed methods over graphs of increasing size, similarly to~\citet{schuetz2022graph}, we present results with respect to the best-of-N metric on graphs with $100$, $1000$, and $10000$ nodes in~\cref{table:scale1,table:scale2,table:scale3}.
Due to the size of our experimental grid and resources needed, the experiments on the largest graphs of size $N=10000$ were restricted to the following settings for the MaxCut problem: we tested all of our $8$ candidate architectures on single random graphs of regularities $d \in 
\{3, 5, 10, 20, 30, 40, 50\}$ with $5$ random initializations. The results show that the discretization methods remain a promising improvement with the scaling of the size of the input.

\begin{table*}[h!]
\caption{Table of results for the maximum cut problem with $N=100$ nodes, w.r.t. the best-of-N metric}
\label{table:scale1}
\resizebox{\textwidth}{!}{%
\begin{tabular}{ccccccccccccccc}
\toprule
 \multirow{2}{*}{MaxCut BoN} & \multicolumn{2}{c}{$d=3$} & \multicolumn{2}{c}{$d=5$} & \multicolumn{2}{c}{$d=10$} & \multicolumn{2}{c}{$d=20$} & \multicolumn{2}{c}{$d=30$} & \multicolumn{2}{c}{$d=40$} & \multicolumn{2}{c}{$d=50$} \\\cmidrule(lr){2-3}\cmidrule(lr){4-5}\cmidrule(lr){6-7}\cmidrule(lr){8-9}\cmidrule(lr){10-11}\cmidrule(lr){12-13}\cmidrule(lr){14-15}
& BoN     & RR\textsubscript{BoN} & BoN     & RR\textsubscript{BoN} & BoN     & RR\textsubscript{BoN} & BoN     & RR\textsubscript{BoN} & BoN     & RR\textsubscript{BoN} & BoN     & RR\textsubscript{BoN} & BoN     & RR\textsubscript{BoN} \\\midrule
\texttt{baseline} & 129.25 & 0.60 & 191.80 & 0.40 & 107.80 & 0.28 & 0.00 & 0.25 & 0.00 & 0.25 & 0.00 & 0.25 & 0.00 & 0.25 \\
\texttt{temp-lin} & 129.35 & 0.64 & 194.40 & 0.61 & 47.00 & 0.21 & 0.00 & 0.25 & 0.00 & 0.25 & 0.00 & 0.25 & 0.00 & 0.25 \\
\texttt{temp-log} & 129.15 & 0.55 & 183.70 & 0.34 & 0.00 & 0.18 & 0.00 & 0.25 & 0.00 & 0.25 & 0.00 & 0.25 & 0.00 & 0.25 \\
\texttt{temp-exp} & \textbf{129.55} & \textbf{0.69} & 192.15 & 0.42 & 77.60 & 0.28 & 0.00 & 0.25 & 0.00 & 0.25 & 0.00 & 0.25 & 0.00 & 0.25 \\
\texttt{bin-ste} & 112.15 & 0.13 & 164.00 & 0.14 & 245.70 & 0.27 & 432.20 & 0.70 & \textbf{631.70} & \textbf{0.88} & 838.70 & 0.88 & \textbf{1017.90} & \textbf{0.89} \\
\texttt{bin-sig} & 119.10 & 0.19 & 146.30 & 0.13 & 240.60 & 0.29 & 432.20 & 0.70 & 627.60 & 0.84 & \textbf{844.30} & \textbf{0.93} & 1015.80 & \textbf{0.89} \\
\texttt{fuzzy-std} & 125.75 & 0.31 & \textbf{196.45} & \textbf{0.81} & \textbf{326.80} & \textbf{0.98} & 0.00 & 0.25 & 0.00 & 0.25 & 0.00 & 0.25 & 0.00 & 0.25 \\
\texttt{fuzzy-luk} & 125.30 & 0.20 & 183.70 & 0.23 & 284.50 & 0.41 & \textbf{442.20} & \textbf{0.73} & 591.50 & 0.51 & 740.30 & 0.38 & 928.10 & 0.47 \\\bottomrule
\end{tabular}%
}
\end{table*}

\begin{table*}[h!]
\caption{Table of results for the maximum cut problem with $N=1000$ nodes, w.r.t. the best-of-N metric}
\label{table:scale2}
\resizebox{\textwidth}{!}{%
\begin{tabular}{ccccccccccccccc}
\toprule
 \multirow{2}{*}{MaxCut BoN} & \multicolumn{2}{c}{$d=3$} & \multicolumn{2}{c}{$d=5$} & \multicolumn{2}{c}{$d=10$} & \multicolumn{2}{c}{$d=20$} & \multicolumn{2}{c}{$d=30$} & \multicolumn{2}{c}{$d=40$} & \multicolumn{2}{c}{$d=50$} \\\cmidrule(lr){2-3}\cmidrule(lr){4-5}\cmidrule(lr){6-7}\cmidrule(lr){8-9}\cmidrule(lr){10-11}\cmidrule(lr){12-13}\cmidrule(lr){14-15}
& BoN     & RR\textsubscript{BoN} & BoN     & RR\textsubscript{BoN} & BoN     & RR\textsubscript{BoN} & BoN     & RR\textsubscript{BoN} & BoN     & RR\textsubscript{BoN} & BoN     & RR\textsubscript{BoN} & BoN     & RR\textsubscript{BoN} \\\midrule
\texttt{baseline} & 1273.90 & 0.36 & 1861.70 & 0.21 & 3144.10 & 0.33 & 0.00 & 0.25 & 0.00 & 0.25 & 0.00 & 0.25 & 0.00 & 0.25 \\
\texttt{temp-lin} & 1277.35 & 0.56 & 1878.60 & 0.38 & 3165.00 & 0.43 & 0.00 & 0.25 & 0.00 & 0.25 & 0.00 & 0.25 & 0.00 & 0.25 \\
\texttt{temp-log} & \textbf{1283.10} & \textbf{0.86} & 1887.70 & 0.46 & 156.20 & 0.13 & 0.00 & 0.25 & 0.00 & 0.25 & 0.00 & 0.25 & 0.00 & 0.25 \\
\texttt{temp-exp} & 1274.25 & 0.36 & 1866.60 & 0.25 & 3153.90 & 0.34 & 0.00 & 0.25 & 0.00 & 0.25 & 0.00 & 0.25 & 0.00 & 0.25 \\
\texttt{bin-ste} & 1062.30 & 0.12 & 1469.05 & 0.12 & 2089.10 & 0.16 & 3987.40 & 0.65 & \textbf{5914.30} & \textbf{0.93} & 7719.40 & 0.95 & \textbf{9699.70} & \textbf{0.98} \\
\texttt{bin-sig} & 1180.65 & 0.15 & 1577.30 & 0.14 & 2064.20 & 0.16 & 3986.70 & 0.64 & 5911.20 & 0.92 & \textbf{7721.90} & \textbf{0.93} & 9699.00 & 0.98 \\
\texttt{fuzzy-std} & 1236.75 & 0.20 & \textbf{1932.90} & \textbf{1.00} & \textbf{3260.90} & \textbf{1.00} & 0.00 & 0.25 & 0.00 & 0.25 & 0.00 & 0.25 & 0.00 & 0.25 \\
\texttt{fuzzy-luk} & 1220.95 & 0.17 & 1698.15 & 0.17 & 2623.30 & 0.20 & \textbf{4122.70} & \textbf{0.80} & 5551.70 & 0.40 & 6967.20 & 0.37 & 8482.10 & 0.33 \\\bottomrule
\end{tabular}%
}
\end{table*}

\begin{table*}[h!]
\centering
\caption{Table of results for the maximum cut problem with $N=10000$ nodes, w.r.t. the best-of-N metric}
\label{table:scale3}
\resizebox{\textwidth}{!}{%
\begin{tabular}{ccccccccccccccc}
\toprule
 \multirow{2}{*}{MaxCut BoN} & \multicolumn{2}{c}{$d=3$} & \multicolumn{2}{c}{$d=5$} & \multicolumn{2}{c}{$d=10$} & \multicolumn{2}{c}{$d=20$} & \multicolumn{2}{c}{$d=30$} & \multicolumn{2}{c}{$d=40$} & \multicolumn{2}{c}{$d=50$} \\\cmidrule(lr){2-3}\cmidrule(lr){4-5}\cmidrule(lr){6-7}\cmidrule(lr){8-9}\cmidrule(lr){10-11}\cmidrule(lr){12-13}\cmidrule(lr){14-15}
& Avg     & RR\textsubscript{Avg} & Avg     & RR\textsubscript{Avg} & Avg     & RR\textsubscript{Avg} & Avg     & RR\textsubscript{Avg} & Avg     & RR\textsubscript{Avg} & Avg     & RR\textsubscript{Avg} & Avg     & RR\textsubscript{Avg} \\\midrule
\texttt{baseline} & 12572 & 0.25 & 18592 & 0.20 & 31272 & 0.25 & 0 & 0.20 & 0 & 0.25 & 0 & 0.25 & 0 & 0.25 \\
\texttt{temp-lin} & 12655 & 0.50 & 18699 & 0.33 & 31538 & 0.50 & 0 & 0.20 & 0 & 0.25 & 0 & 0.25 & 0 & 0.25 \\
\texttt{temp-log} & \textbf{12779} & \textbf{1.00} & 18774 & 0.50 & 31216 & 0.20 & 0 & 0.20 & 0 & 0.25 & 0 & 0.25 & 0 & 0.25 \\
\texttt{temp-exp} & 12592 & 0.33 & 18619 & 0.25 & 31414 & 0.33 & 0 & 0.20 & 0 & 0.25 & 0 & 0.25 & 0 & 0.25 \\
\texttt{bin-ste} & 10120 & 0.12 & 13862 & 0.12 & 18980 & 0.14 & 37666 & 0.33 & \textbf{55284} & \textbf{1.00} & \textbf{73722} & \textbf{1.00} & \textbf{92576} & \textbf{1.00} \\
\texttt{bin-sig} & 11637 & 0.14 & 15252 & 0.14 & 18978 & 0.12 & 37666 & 0.33 & \textbf{55284} & \textbf{1.00} & \textbf{73722} & \textbf{1.00} & \textbf{92576} & \textbf{1.00} \\
\texttt{fuzzy-std} & 12270 & 0.20 & \textbf{19279} & \textbf{1.00} & \textbf{32538} & \textbf{1.00} & \textbf{57174} & \textbf{1.00} & 0 & 0.25 & 0 & 0.25 & 0 & 0.25 \\
\texttt{fuzzy-luk} & 12191 & 0.17 & 16719 & 0.17 & 26170 & 0.17 & 39714 & 0.50 & 53486 & 0.33 & 66974 & 0.33 & 80288 & 0.33 \\\bottomrule
\end{tabular}%
}
\end{table*}

\section{Results by Average Performance}
\label{app:resavg}
The main text presents results based on the best-of-N (BoN) aggregation, as it is the preferred one with respect to the parallelization of the training~\citep{schuetz2022combinatorial}. For completeness, see the results for $N \in \{100, 1000\}$ based on averaging (Avg) metric in~\cref{table:5,table:6,table:7,table:8}.
The averaging metric requires adjustments for non-feasible solutions (which are possible only for MIS). While in the BoN metrics we could disregard such results as long as at least one of the seeds for each setting resulted in a feasible solution, for Avg, infeasible results should logically affect the final arithmetic mean. We nullified such results, with the intuition that the bitstring of zeros represents the trivial result of empty maximal sets and cuts.

\begin{table*}[h!]
\centering
\vspace{0.5em}
\caption{Table of results for the maximum cut problem with $N=100$ nodes, w.r.t. the average metric}
\label{table:5}
\resizebox{\textwidth}{!}{%
\begin{tabular}{ccccccccccccccc}
\toprule
 \multirow{2}{*}{MaxCut Avg} & \multicolumn{2}{c}{$d=3$} & \multicolumn{2}{c}{$d=5$} & \multicolumn{2}{c}{$d=10$} & \multicolumn{2}{c}{$d=20$} & \multicolumn{2}{c}{$d=30$} & \multicolumn{2}{c}{$d=40$} & \multicolumn{2}{c}{$d=50$} \\\cmidrule(lr){2-3}\cmidrule(lr){4-5}\cmidrule(lr){6-7}\cmidrule(lr){8-9}\cmidrule(lr){10-11}\cmidrule(lr){12-13}\cmidrule(lr){14-15}
& Avg     & RR\textsubscript{Avg} & Avg     & RR\textsubscript{Avg} & Avg     & RR\textsubscript{Avg} & Avg     & RR\textsubscript{Avg} & Avg     & RR\textsubscript{Avg} & Avg     & RR\textsubscript{Avg} & Avg     & RR\textsubscript{Avg} \\\midrule
\texttt{baseline} & 125.22 & 0.46 & 184.24 & 0.32 & 24.48 & 0.20 & 0.00 & 0.25 & 0.00 & 0.25 & 0.00 & 0.25 & 0.00 & 0.25 \\
\texttt{temp-lin} & \textbf{125.96} & \textbf{0.80} & 184.50 & 0.50 & 9.40 & 0.18 & 0.00 & 0.25 & 0.00 & 0.25 & 0.00 & 0.25 & 0.00 & 0.25 \\
\texttt{temp-log} & 83.72 & 0.15 & 72.76 & 0.14 & 0.00 & 0.18 & 0.00 & 0.25 & 0.00 & 0.25 & 0.00 & 0.25 & 0.00 & 0.25 \\
\texttt{temp-exp} & 125.57 & 0.67 & 184.98 & 0.37 & 18.44 & 0.20 & 0.00 & 0.25 & 0.00 & 0.25 & 0.00 & 0.25 & 0.00 & 0.25 \\
\texttt{bin-ste} & 80.91 & 0.14 & 110.88 & 0.16 & 124.88 & 0.33 & 163.56 & 0.62 & \textbf{237.68} & 0.80 & 316.10 & 0.88 & 384.32 & 0.89 \\
\texttt{bin-sig} & 80.11 & 0.14 & 78.79 & 0.13 & 111.04 & 0.28 & 163.84 & 0.62 & 237.06 & \textbf{0.82} & \textbf{316.58} & \textbf{0.90} & \textbf{385.06} & \textbf{0.92} \\
\texttt{fuzzy-std} & 121.09 & 0.26 & \textbf{190.92} & \textbf{0.93} & \textbf{315.96} & \textbf{1.00} & 0.00 & 0.25 & 0.00 & 0.25 & 0.00 & 0.25 & 0.00 & 0.25 \\
\texttt{fuzzy-luk} & 118.93 & 0.22 & 138.99 & 0.20 & 186.20 & 0.50 & \textbf{195.70} & \textbf{0.83} & 222.86 & 0.57 & 268.28 & 0.37 & 335.08 & 0.41 \\\bottomrule
\end{tabular}%
}
\end{table*}
\begin{table*}[h!]
\centering
\vspace{0.5em}
\caption{Table of results for the maximum cut problem with $N=1000$ nodes, w.r.t. the average metric}
\label{table:6}
\resizebox{\textwidth}{!}{%
\begin{tabular}{ccccccccccccccc}
\toprule
 \multirow{2}{*}{MaxCut Avg} & \multicolumn{2}{c}{$d=3$} & \multicolumn{2}{c}{$d=5$} & \multicolumn{2}{c}{$d=10$} & \multicolumn{2}{c}{$d=20$} & \multicolumn{2}{c}{$d=30$} & \multicolumn{2}{c}{$d=40$} & \multicolumn{2}{c}{$d=50$} \\\cmidrule(lr){2-3}\cmidrule(lr){4-5}\cmidrule(lr){6-7}\cmidrule(lr){8-9}\cmidrule(lr){10-11}\cmidrule(lr){12-13}\cmidrule(lr){14-15}
& Avg     & RR\textsubscript{Avg} & Avg     & RR\textsubscript{Avg} & Avg     & RR\textsubscript{Avg} & Avg     & RR\textsubscript{Avg} & Avg     & RR\textsubscript{Avg} & Avg     & RR\textsubscript{Avg} & Avg     & RR\textsubscript{Avg} \\\midrule
\texttt{baseline} & 1259.17 & 0.27 & 1838.59 & 0.20 & 3091.92 & 0.28 & 0.00 & 0.25 & 0.00 & 0.25 & 0.00 & 0.25 & 0.00 & 0.25 \\
\texttt{temp-lin} & 1263.62 & 0.52 & 1855.19 & 0.37 & 3119.10 & 0.46 & 0.00 & 0.25 & 0.00 & 0.25 & 0.00 & 0.25 & 0.00 & 0.25 \\
\texttt{temp-log} & \textbf{1271.26} & \textbf{0.98} & 1865.86 & 0.48 & 31.24 & 0.12 & 0.00 & 0.25 & 0.00 & 0.25 & 0.00 & 0.25 & 0.00 & 0.25 \\
\texttt{temp-exp} & 1260.34 & 0.33 & 1844.28 & 0.25 & 3104.32 & 0.36 & 0.00 & 0.25 & 0.00 & 0.25 & 0.00 & 0.25 & 0.00 & 0.25 \\
\texttt{sign-ste} & 972.45 & 0.12 & 1366.95 & 0.12 & 1865.12 & 0.16 & 2756.34 & 0.48 & 3707.94 & 0.48 & 4706.62 & 0.49 & 5727.72 & 0.48 \\
\texttt{sign-sig} & 1133.53 & 0.14 & 1447.32 & 0.14 & 1622.10 & 0.15 & 2756.16 & 0.48 & 3706.60 & 0.48 & 4706.92 & 0.48 & 5737.10 & 0.48 \\
\texttt{fuzzy-std} & 1221.92 & 0.20 & \textbf{1917.13} & \textbf{1.00} & \textbf{3228.98} & \textbf{1.00} & 0.00 & 0.25 & 0.00 & 0.25 & 0.00 & 0.25 & 0.00 & 0.25 \\
\texttt{fuzzy-luk} & 1197.24 & 0.17 & 1649.62 & 0.17 & 2459.58 & 0.20 & \textbf{3178.60} & \textbf{1.00} & \textbf{4162.54} & \textbf{1.00} & \textbf{5169.68} & \textbf{1.00} & \textbf{6382.56} & \textbf{1.00} \\\bottomrule
\end{tabular}%
}
\end{table*}

\begin{table*}[h!]
\centering
\vspace{0.5em}
\caption{Table of results for the maximum independent set problem with $N=100$ nodes, w.r.t. the average metric}
\label{table:7}
\resizebox{\textwidth}{!}{%
\begin{tabular}{ccccccccccccccc}
\toprule
 \multirow{2}{*}{MIS Avg} & \multicolumn{2}{c}{$d=3$} & \multicolumn{2}{c}{$d=5$} & \multicolumn{2}{c}{$d=10$} & \multicolumn{2}{c}{$d=20$} & \multicolumn{2}{c}{$d=30$} & \multicolumn{2}{c}{$d=40$} & \multicolumn{2}{c}{$d=50$} \\\cmidrule(lr){2-3}\cmidrule(lr){4-5}\cmidrule(lr){6-7}\cmidrule(lr){8-9}\cmidrule(lr){10-11}\cmidrule(lr){12-13}\cmidrule(lr){14-15}
& Avg     & RR\textsubscript{Avg} & Avg     & RR\textsubscript{Avg} & Avg     & RR\textsubscript{Avg} & Avg     & RR\textsubscript{Avg} & Avg     & RR\textsubscript{Avg} & Avg     & RR\textsubscript{Avg} & Avg     & RR\textsubscript{Avg} \\\midrule
\texttt{baseline} & 41.53 & 0.50 & 10.48 & 0.29 & 0.00 & 0.25 & 0.00 & 0.25 & 0.00 & 0.26 & 0.00 & 0.25 & 0.00 & 0.25 \\
\texttt{temp-lin} & \textbf{41.86} & \textbf{0.98} & 6.21 & 0.22 & 0.00 & 0.25 & 0.00 & 0.25 & 0.00 & 0.26 & 0.00 & 0.25 & 0.00 & 0.25 \\
\texttt{temp-log} & 26.44 & 0.18 & 0.73 & 0.15 & 0.00 & 0.25 & 0.00 & 0.25 & 0.00 & 0.26 & 0.00 & 0.25 & 0.00 & 0.25 \\
\texttt{temp-exp} & 41.64 & 0.67 & 9.53 & 0.27 & 0.00 & 0.25 & 0.00 & 0.25 & 0.00 & 0.26 & 0.00 & 0.25 & 0.00 & 0.25 \\
\texttt{bin-ste} & 25.32 & 0.13 & 12.92 & 0.24 & 4.52 & 0.54 & 1.94 & \textbf{0.88} & \textbf{1.20} & \textbf{0.78} & 1.01 & \textbf{0.93} & 0.76 & \textbf{0.88} \\
\texttt{bin-sig} & 21.47 & 0.14 & 9.26 & 0.18 & 4.50 & 0.53 & 1.94 & \textbf{0.88} & \textbf{1.20} & \textbf{0.78} & 1.01 & \textbf{0.93} & 0.76 & \textbf{0.88} \\
\texttt{fuzzy-std} & 40.53 & 0.27 & \textbf{33.18} & \textbf{1.00} & 0.00 & 0.25 & 0.00 & 0.25 & 0.00 & 0.26 & 0.00 & 0.25 & 0.00 & 0.25 \\
\texttt{fuzzy-luk} & 38.61 & 0.19 & 25.03 & 0.46 & \textbf{8.73} & \textbf{0.93} & \textbf{2.14} & 0.60 & 1.15 & 0.70 & \textbf{1.18} & 0.53 & \textbf{0.96} & 0.87 \\\bottomrule
\end{tabular}%
}
\end{table*}
\begin{table*}[h!]
\centering
\caption{Table of results for the maximum independent set problem with $N=1000$ nodes, w.r.t. the average metric}
\label{table:8}
\resizebox{\textwidth}{!}{%
\begin{tabular}{ccccccccccccccc}
\toprule
 \multirow{2}{*}{MIS Avg} & \multicolumn{2}{c}{$d=3$} & \multicolumn{2}{c}{$d=5$} & \multicolumn{2}{c}{$d=10$} & \multicolumn{2}{c}{$d=20$} & \multicolumn{2}{c}{$d=30$} & \multicolumn{2}{c}{$d=40$} & \multicolumn{2}{c}{$d=50$} \\\cmidrule(lr){2-3}\cmidrule(lr){4-5}\cmidrule(lr){6-7}\cmidrule(lr){8-9}\cmidrule(lr){10-11}\cmidrule(lr){12-13}\cmidrule(lr){14-15}
& Avg     & RR\textsubscript{Avg} & Avg     & RR\textsubscript{Avg} & Avg     & RR\textsubscript{Avg} & Avg     & RR\textsubscript{Avg} & Avg     & RR\textsubscript{Avg} & Avg     & RR\textsubscript{Avg} & Avg     & RR\textsubscript{Avg} \\\midrule
\texttt{baseline} & 417.98 & 0.24 & 340.44 & 0.35 & 0.00 & 0.25 & 0.00 & 0.40 & 0.00 & 0.33 & 0.00 & 0.31 & 0.00 & 0.26 \\
\texttt{temp-lin} & 419.59 & 0.48 & \textbf{344.52} & \textbf{0.74} & 0.00 & 0.25 & 0.00 & 0.40 & 0.00 & 0.33 & 0.00 & 0.31 & 0.00 & 0.26 \\
\texttt{temp-log} & \textbf{423.28} & \textbf{0.94} & 244.86 & 0.32 & 0.00 & 0.25 & 0.00 & 0.40 & 0.00 & 0.33 & 0.00 & 0.31 & 0.00 & 0.26 \\
\texttt{temp-exp} & 418.60 & 0.32 & 341.81 & 0.50 & 0.00 & 0.25 & 0.00 & 0.40 & 0.00 & 0.33 & 0.00 & 0.31 & 0.00 & 0.26 \\
\texttt{bin-ste} & 303.64 & 0.14 & 220.69 & 0.15 & 48.92 & 0.50 & 23.72 & 0.50 & \textbf{15.16} & 0.49 & \textbf{12.08} & 0.55 & 8.89 & 0.62 \\
\texttt{bin-sig} & 360.98 & 0.17 & 104.64 & 0.14 & 48.28 & 0.49 & 23.84 & 0.50 & 15.08 & 0.50 & \textbf{12.08} & 0.55 & \textbf{8.91} & 0.65 \\
\texttt{fuzzy-std} & 408.13 & 0.33 & 332.13 & 0.38 & 0.00 & 0.25 & 0.00 & 0.40 & 0.00 & 0.33 & 0.00 & 0.31 & 0.00 & 0.26 \\
\texttt{fuzzy-luk} & 393.67 & 0.14 & 298.23 & 0.16 & \textbf{170.51} & \textbf{1.00} & \textbf{46.35} & \textbf{1.00} & 13.89 & \textbf{1.00} & 10.89 & \textbf{0.97} & 8.67 & \textbf{0.80} \\\bottomrule
\end{tabular}%
}
\end{table*}

\section{Comparison to Solvers}
Similarly to previous work, the results on smaller instances have been compared with solvers. Notably, for MaxCut, the Goemans-Williamson algorithm~\citep{goemans1995improved} has been used.
The solver can, however, handle only graphs with a couple of hundred nodes in a reasonable time and with reasonable memory demands~\citep{schuetz2022combinatorial}.
On data with $N=100$ nodes we have observed: (i) comparable results on the low-regularity graphs, retrieving $0.97$ to $0.98$ of the solver's prediction value with our improved PI-GNN approach; (ii) lower results on the high-regularity graphs, retrieving $0.71$ to $0.72$ of the solver's prediction value with our improved PI-GNN approach.
Thus, while our approach surpasses the original PI-GNNs proposal~\citep{schuetz2022combinatorial} and, unlike classical solvers, remains efficient on larger instances, improving PI-GNNs' performance on small graphs, over classical solvers, remains open for further exploration.

\section{Regularization Ablation}
\label{app:ablation}
As an intuitive alternative to soft discretization, a regularization term can possibly be added to the QUBO loss, penalizing predictions based on their distance from a discrete value:
\begin{equation}
{H^\prime_{Q_\cdot}}(\theta, x) = {H_{Q_\cdot}}(\theta, x) + \alpha \cdot P(x),
\end{equation}
with the constant $\alpha$ controlling the ratio of importance between QUBO loss and the regularization penalty.

Concretely, two regularizations were considered: the \textit{L1 penalty}: $P(x) = \sum_{i=1}^N |x_i|$,
and \textit{binary entropy}: $P(x) = \sum_{i=1}^N [-x_i \cdot log_2(x_i) - (1 - x_i) \cdot log_2(1 - x_i)]$.
The former is more directed towards sparsity, as it encourages values close to zero.
The latter is suited directly for binary discretization, being symmetric for $0$ and $1$.

The experiments (with multiple values of $\alpha$) showed a similar or worse performance compared to the original PI-GNNs (\texttt{baseline}). The \textit{binary entropy} performed slightly better than \textit{L1 penalty}.
However, neither of the regularized losses showed sufficient performance to include them in the proposed portfolio of methods.

\section{Hyperparameters}
Training and GNN hyperparameters are disclosed in~\cref{table:hyper}. The settings, following the methodology of~\citet{schuetz2022combinatorial}, were shared over all the experiments.
\label{app:hyperparams}
\begin{table}[h!]
\centering
\caption{Overview of Deep Learning Hyperparameters}
\label{table:hyper}
\begin{tabular}{@{}lll@{}}
\toprule
\textbf{Hyperparameter}          & \textbf{Description}                                   & \textbf{Value} \\ 
\midrule
Number of Layers                 & Total number of graph convolutional layers            & $2$              \\
Graph Convolutional Layer        & Type of graph convolutional operator                & \citet{kipf2016semi}      \\
Graph size             & Number of graph vertices             & $N \in \{100, 1000, 10000\}$             \\
Input Embedding Size             & Dimensionality of input node embeddings             & $d_0 = \sqrt{N}$             \\
Hidden Dimensions                & Dimensionality of hidden layer representations        & $d_1 = int(d_0 / 2)$             \\
Intermediate Activation Function & Non-linear activation function for intermediate layers & ReLU           \\
Final Activation Function        & Non-linear activation function for output layer       & Sigmoid / Sigmoid\textsubscript{temp} / Step        \\
Loss function        & Unsupervised loss encoding the CO problem       & QUBO / QUBO\textsubscript{S} / QUBO\textsubscript{L}        \\
Optimizer                        & Optimization algorithm                                & Adam~\citep{kingma2014adam}      \\
Learning Rate                    & Step size for optimizer updates                       & $10^{-4}$           \\
Maximum Epochs                   & Nr. of training epochs if early stopping not triggered                      & $10^5$            \\
Early Stopping Tolerance         & Minimum improvement required to continue training     & $10^{-4}$           \\
Early Stopping Patience          & Epochs with no improvement before stopping            & $10^3$             \\
\bottomrule
\end{tabular}
\end{table}

\section{Heatmaps of Node Assignment Dynamics}
\label{app:dynamics}
\begin{figure}[h!]
    \centering
    \includegraphics[width=\linewidth]{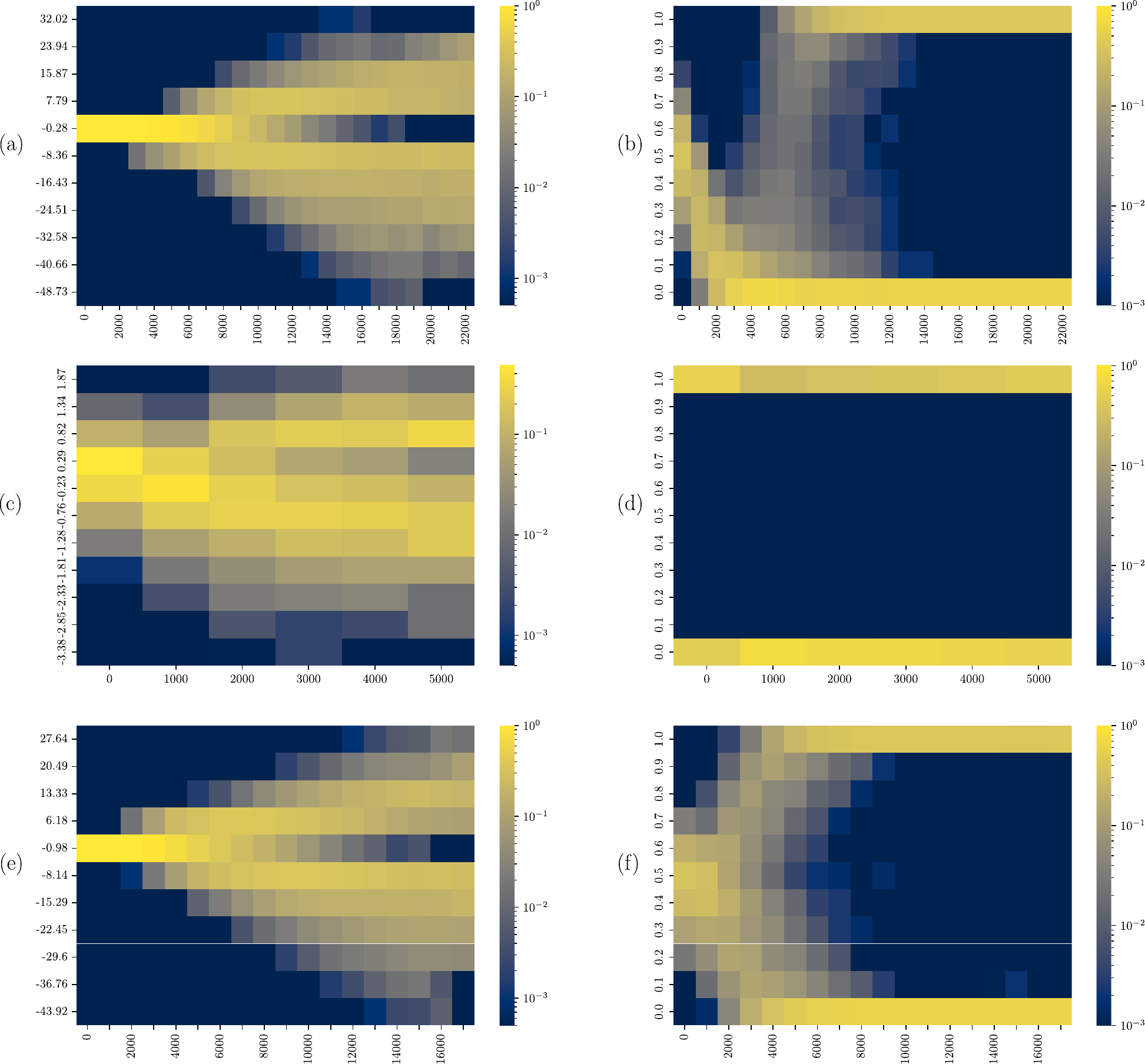}
    \caption{Distributions of GNN's pre-activation ($a^{pre}$) and post-activation ($a^{post}$) values on $5$-regular \textit{MaxCut} graphs: (a) \texttt{baseline} pre-activations, (b) \texttt{baseline} post-activations, (c) \texttt{sign-ste} pre-activations, (d) \texttt{sign-ste} post-activations, (e) \texttt{fuzzy-luk} pre-activations, (f) \texttt{fuzzy-luk} post-activations.}
    \label{fig:values-nxn-maxcut-5}
\end{figure}

\begin{figure}[h!]
    \centering
    \includegraphics[width=\linewidth]{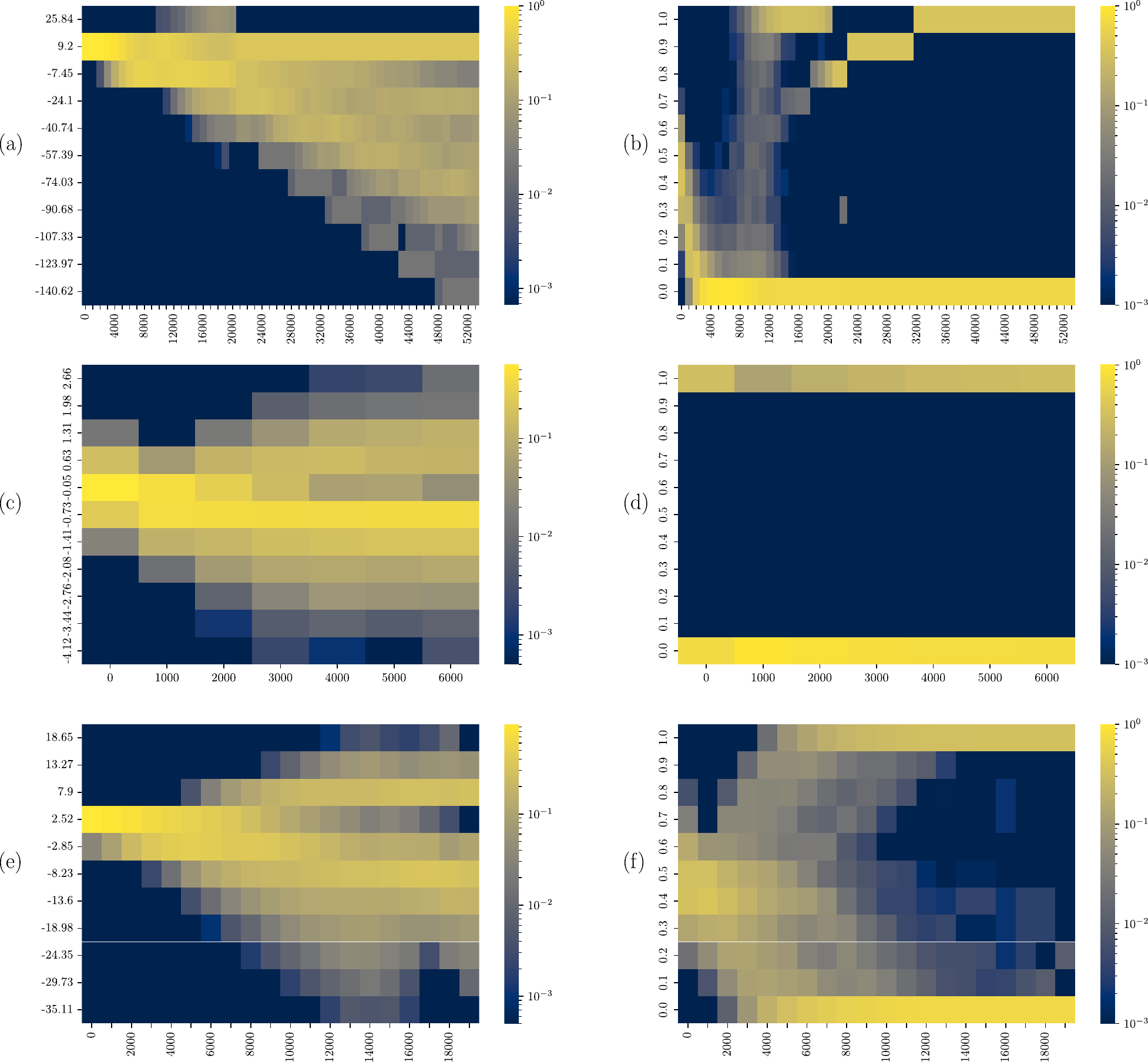}
    \caption{Distributions of GNN's pre-activation ($a^{pre}$) and post-activation ($a^{post}$) values on $5$-regular \textit{MIS} graphs: (a) \texttt{baseline} pre-activations, (b) \texttt{baseline} post-activations, (c) \texttt{sign-ste} pre-activations, (d) \texttt{sign-ste} post-activations, (e) \texttt{fuzzy-luk} pre-activations, (f) \texttt{fuzzy-luk} post-activations.}
    \label{fig:values-nxn-mis-5}
\end{figure}

Further heatmaps depicting the training dynamics for $5$-regular MaxCut and MIS graphs are in~\cref{fig:values-nxn-maxcut-5,fig:values-nxn-mis-5}.
Similarly to graphs in the main text, we can see the phase transition in the upper right corners and improved training dynamics in the second and third rows.


\end{document}